\documentclass{article}

\usepackage{iclr2027_conference,times}

\usepackage[utf8]{inputenc} 
\usepackage[T1]{fontenc}    
\PassOptionsToPackage{hyphens}{url}
\usepackage{hyperref}       
\hypersetup{
  pdftitle={InfraMind: Infrastructure-Aware Multi-Agent Orchestration},
  pdfauthor={Ahasan Kabir, Mengxin Zheng, Qian Lou},
  pdfsubject={Preprint},
  pdfkeywords={}
}
\usepackage{url}            
\usepackage{xurl}           
\usepackage{booktabs}       
\usepackage{amsfonts}       
\usepackage{amsmath}        
\usepackage{amssymb}        
\usepackage{amsthm}
\usepackage{algorithm}
\usepackage{algpseudocode}
\usepackage{microtype}      
\usepackage{xcolor}         
\usepackage{graphicx}       
\usepackage{wrapfig}        
\usepackage{multirow}       
\usepackage{pifont}         
\usepackage{placeins}        
\usepackage{needspace}       

\AddToHook{env/table/begin}{\setlength{\belowcaptionskip}{6pt}}
\AddToHook{env/wraptable/begin}{\setlength{\belowcaptionskip}{6pt}}

\newcommand{\sys}{\textsc{InfraMind}}
\newtheorem{proposition}{Proposition}

\title{\sys{}: Infrastructure-Aware \\ Multi-Agent Orchestration}

\author{Ahasan Kabir, Mengxin Zheng \& Qian Lou \\
University of Central Florida \\
\texttt{\{ahasan.kabir, mengxin.zheng, qian.lou\}@ucf.edu}
}

\iclrfinalcopy

\begin{document}

\maketitle
\lhead{Preprint}

\begin{abstract}
Multi-agent LLM systems spend additional computation on reasoning and collaboration, but shared serving queues can prevent that computation from producing answers before a deadline. This creates a coupled decision: the reasoning workflow determines the load, while changing load determines which reasoning choices remain affordable. We introduce \sys{}, a hierarchical policy that adapts workflow planning and execution to serving conditions. At arrival, a planner selects the collaboration graph using task information, load and the time budget. Before each agent call, an executor jointly selects the model and reasoning strategy using refreshed telemetry and remaining time. A shared latency penalty connects the two policies. Our analysis characterizes why observing load matters when congestion reverses the preferred computation. Across five benchmarks on a shared five-model deployment, \sys{} improves low-load accuracy by up to $7.6$ percentage points over the strongest evaluated baseline. On 500 MATH questions with matched EDF scheduling and per-request deadlines of 10--300\,s, it increases correct, on-time answers from $20.4\%$ to $33.8\%$. With 15 models and Azure timings replayed on the same benchmark and deadline tiers, this fraction rises from $11.6\%$ to $54.8\%$ over MasRouter. These results support adapting reasoning decisions to the serving capacity available throughout a workflow.
\end{abstract}

\section{Introduction}
\label{sec:intro}

Multi-agent LLM systems improve answers through debate, review and sequential reasoning~\citep{wu2024autogen, hong2023metagpt, li2023camel}. Learning which models should collaborate, in which roles and topology, is therefore central to orchestration~\citep{wang2406mixture, zhuge2024gptswarm, yue2025masrouter}. On shared infrastructure, however, more reasoning helps only if it finishes in time. A model that is fast in isolation can become the slowest choice once its queue fills, consuming time needed downstream. The practical challenge is to turn available computation into \emph{correct answers within the available time}.

Task-driven orchestration leaves a crucial variable outside this decision: the \emph{runtime state} of the serving infrastructure. We call this disconnect \emph{infrastructure blindness}. It matters when heterogeneous open-weight models~\citep{grattafiori2024llama, guo2025deepseek} share GPU capacity through engines such as vLLM~\citep{kwon2023efficient} and SGLang~\citep{zheng2024sglang}. Bursty arrivals and variable request lengths make capacity change during a workflow~\citep{xiang2026servegen, wang2025burstgpt}; dependencies between calls compound the resulting waits~\citep{lin2024parrot, luo2026agentix}. If orchestration cannot respond, a deployment can simultaneously have overloaded preferred models, idle alternatives and missed deadlines. Additional model capacity alone does not resolve this mismatch.

\begin{figure}[t]
  \centering
  \includegraphics[width=\textwidth]{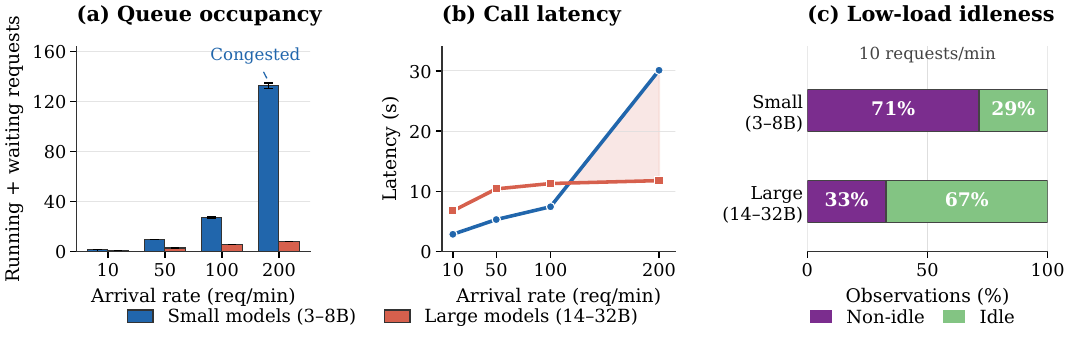}
  \caption{Load-agnostic routing concentrates work while alternatives have spare capacity (MasRouter, MATH, Poisson arrivals). \textbf{(a)}~Mean running + waiting requests at sampled calls; error bars show standard errors. \textbf{(b)}~Mean call latency by model group. \textbf{(c)}~Idle and non-idle observations at 10 requests/min; idle means zero running + waiting requests.}
  \label{fig:motivation}
\end{figure}

Figure~\ref{fig:motivation} profiles MasRouter~\citep{yue2025masrouter} on five shared models under Poisson load. At 200 requests/min, small models average 133 running or waiting requests versus 8 for larger models, with call latencies of 30.1 versus 11.8\,s. At 10 requests/min, 67\% of large-model telemetry observations record no running or waiting requests. Despite this changing capacity, a twentyfold increase in offered load changes every model's routing share by at most $1.11$ percentage points (Appendix~\ref{app:invariance}). The same task-based preferences persist across operating conditions that call for different computation.

The difficulty is that \emph{the reasoning policy also creates the load it must handle}. Choosing more agents increases the work admitted to the pool; choosing a model and reasoning depth changes both answer quality and service demand; waiting at one step reduces the budget for every later step. These decisions operate at different timescales, and the load observed when a graph is planned can change before its calls execute. A shortest-queue rule cannot judge whether a model can solve the task, while a quality-only router cannot judge whether its preferred computation is affordable. Queue ordering can prioritize already-chosen calls, but cannot revise their model or reasoning strategy. The unresolved decision is therefore how to allocate the reasoning itself as task needs, serving capacity and remaining time interact.

\begin{figure}[t]
  \centering
  \includegraphics[width=\textwidth]{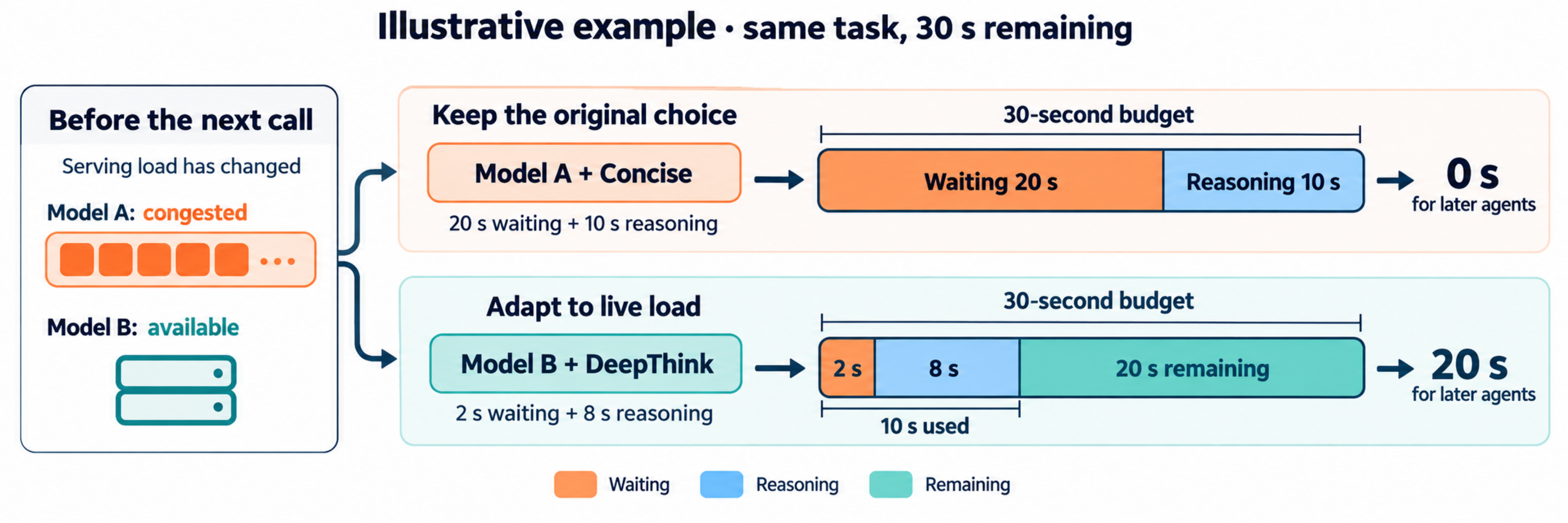}
  \caption{Why adaptation helps: with 30\,s remaining, switching from congested Model A + Concise to available Model B + DeepThink leaves 20\,s for downstream agents. Times are illustrative, not measured; bars are schematic.}
  \label{fig:overview}
\end{figure}
We propose \sys{}, a hierarchical policy for \emph{allocating multi-agent reasoning under live serving conditions}. Each design choice addresses a challenge above. More agents create competing work, so the planner selects topology, agent count and roles using the initial load and budget. Queues change during execution, so the executor refreshes telemetry before each call. Waiting reduces downstream time, so model and reasoning strategy (Flash / Concise / DeepThink) are chosen jointly using the remaining budget. A common latency penalty connects these decisions. Bypassing a congested model can then release time for deeper reasoning elsewhere (Figure~\ref{fig:overview}).

The planner builds on MasRouter's controller~\citep{yue2025masrouter}; FiLM, EDF and the learning algorithms are established components. The contribution is a hierarchical formulation that connects the work created at arrival to execution-time model--strategy decisions through a shared latency penalty.

\textbf{Contributions.} (1)~We formulate capacity-dependent reasoning allocation and characterize the utility gap when unobserved load reverses the preferred action (Proposition~\ref{prop:blindness}). (2)~We develop the coupled planner and executor (Algorithm~\ref{alg:inframind}). (3)~We evaluate five benchmarks and test learned selection and telemetry with paired ablations. On 500 MATH questions at 50 requests/min with 10--300\,s deadline tiers, correct, on-time answers rise from $20.4\%$ to $33.8\%$ under matched EDF and from $11.6\%$ to $54.8\%$ with 15 models and replayed Azure timings (\S\ref{sec:experiments}).

\section{Related Work}
\label{sec:related}

\paragraph{Multi-agent orchestration decides the work.}
MoA~\citep{wang2406mixture} ensembles the model pool; GPTSwarm~\citep{zhuge2024gptswarm} learns a collaboration graph; MasRouter~\citep{yue2025masrouter} selects topology, roles and models from query semantics. Further work adapts agent teams~\citep{liu2024dylan}, prunes communication~\citep{zhang2025agentprune}, searches workflows~\citep{zhang2025aflow, hu2025automated}, or revises topology during execution~\citep{lu2026dytopo, xu2026evomas}. These methods use task information and intermediate outputs, but do not condition model selection on live serving telemetry.

\paragraph{Serving systems decide where and when.}
Serving systems optimize cache management~\citep{kwon2023efficient, zheng2024sglang}, batching~\citep{yu2022orca, agrawal2024taming}, phase placement and migration~\citep{zhong2024distserve, sun2024llumnix}, and admission or instance mapping~\citep{qin2025mooncake, yuan2026dualmap}. Parrot~\citep{lin2024parrot}, Agentix~\citep{luo2026agentix} and Kairos~\citep{chen2025kairos} additionally exploit workflow context. These systems improve execution of application-selected calls. Single-query routers such as RouteLLM~\citep{ong2024routellm} and TREACLE~\citep{NEURIPS2024_a6deba3b} instead adapt model use to quality and cost; R2-Router~\citep{xue2026r2} jointly selects model and output length.

\paragraph{Our methodological distinction.}
\sys{} conditions the reasoning graph and its execution on serving state at two timescales. It extends MasRouter's planning foundation with load-conditioned graph selection and stepwise model--reasoning decisions. Relative to R2-Router, these choices respond to live contention throughout a multi-agent workflow. Relative to Agentix and Kairos, serving conditions change the requested computation itself. A shared latency penalty couples planning to execution (\S\ref{sec:training}).

\section{Problem Formulation}
\label{sec:problem}

Orchestration decides at two timescales: \emph{what reasoning structure to use} at query arrival, and \emph{how to execute each step} as queues and remaining time change. We formalize this as a hierarchical Constrained Markov Decision Process (CMDP)~\citep{altman2021constrained}. The planner samples a graph specification $g=(\tau,K,\mathbf{r})$, comprising topology, agent count and roles, from query semantics and the initial system summary $\mathbf{z}_0=[b_0,\mathbf{d}^{\text{queue}},\mathbf{d}^{\text{e2e}},\mathbf{d}^{\text{kv}}]$. The policy then operates as
\begin{equation}
  \underbrace{g\sim\pi_{\text{plan}}(\cdot\mid\mathbf{e}_q,\mathbf{z}_0)}_{\text{once per query}},
  \qquad
  \underbrace{a_k=(m_k,\sigma_k)\sim\pi_{\text{exec}}(\cdot\mid s_k)}_{\text{at each agent step}}.
  \label{eq:hierarchy}
\end{equation}
Each executor state is refreshed after earlier decisions have affected the workload. The graph determines which roles must execute, while the executor determines the computation used to realize them. Throughout, a query's \emph{budget} is a latency deadline in seconds.

\paragraph{State.} Consider $N$ LLM services $\mathcal{M} = \{m_1, \ldots, m_N\}$ on shared GPUs and prompting strategies $\mathcal{S} = \{\text{Flash}, \text{Concise}, \text{DeepThink}\}$. A query $q$ arrives with a time budget $\beta$. At each agent step $k$, the executor observes:
\begin{equation}
  s_k = \big[\underbrace{\mathbf{e}_q,\; \mathbf{e}_{r_k}}_{\text{what to solve}},\;\; \underbrace{b_k}_{\text{time left}},\;\; \underbrace{\mathbf{d}^{\text{queue}},\; \mathbf{d}^{\text{e2e}},\; \mathbf{d}^{\text{kv}}}_{\text{system load}}\big]
  \label{eq:state}
\end{equation}
where $\mathbf{e}_q, \mathbf{e}_{r_k}$ are Sentence-BERT~\citep{reimers2019sentence} query and role embeddings, $b_k$ is the normalized budget remaining along the agent's dependency path, and $\mathbf{d}^{\text{queue}}, \mathbf{d}^{\text{e2e}}, \mathbf{d}^{\text{kv}} \in \mathbb{R}^N$ are per-model queue depth, end-to-end latency and KV cache utilization polled from vLLM's \texttt{/metrics} endpoint.

We encode the pair $a_k=(m_k,\sigma_k)$ as an index $j_k\in\{0,\ldots,N|\mathcal{S}|-1\}$, with model index $\lfloor j_k/|\mathcal{S}|\rfloor$ and strategy index $j_k\bmod|\mathcal{S}|$.

\paragraph{Objective.} Let $Y\in\{0,1\}$ denote correctness of the final workflow answer and $L$ its elapsed latency. We seek $\pi^*=\arg\max_\pi\mathbb{E}_\pi[Y]$ subject to $\mathbb{E}_\pi[L/\beta]\leq1$. The Lagrangian is $\mathcal{L}(\pi,\lambda)=\mathbb{E}_\pi[Y-\lambda L/\beta]+\lambda$. One multiplier prices time for both graph planning and step execution; the dual update (Eq.~\ref{eq:dual}) raises that price when workflows overspend and lowers it when they have slack. Training uses this expected-cost constraint, while evaluation measures deadline attainment per request.

\subsection{The Cost of Infrastructure Blindness}
\label{sec:theory}

We isolate the value of serving observations at one decision. Fix the task, role, budget and prior execution history $x$, a multiplier $\lambda\geq0$, and a continuation policy and scheduler. Let $Z$ denote the serving state at that decision, and let $v_\lambda(x,z,a)$ be the expected final utility $Y-\lambda L/\beta$ when action $a$ is taken in state $z$ and the fixed continuation follows. Assume a finite action set and integrable utilities; both decision rules below face the same conditional distribution of $Z$ given $x$.

\begin{proposition}[Utility gap from unobserved load]
\label{prop:blindness}
The optimal state-conditioned and state-blind decision values satisfy
\begin{equation}
  V_{\rm live}=\mathbb{E}_{Z\mid x}\!\left[\max_a v_\lambda(x,Z,a)\right]
  \;\geq\;
  V_{\rm blind}=\max_a\mathbb{E}_{Z\mid x}[v_\lambda(x,Z,a)].
  \label{eq:information}
\end{equation}
For two actions $a_0,a_1$ and two load states $z_0,z_1$ with probabilities $p,1-p$, suppose the preferred action reverses: $\delta_0=v_\lambda(x,z_0,a_0)-v_\lambda(x,z_0,a_1)>0$ and $\delta_1=v_\lambda(x,z_1,a_1)-v_\lambda(x,z_1,a_0)>0$. Then
\begin{equation}
  V_{\rm live}-V_{\rm blind}
  =\min\{p\delta_0,(1-p)\delta_1\}>0,\qquad 0<p<1.
  \label{eq:blindness_gap}
\end{equation}
The bound includes randomized state-blind decisions.
\end{proposition}
\begin{proof}
Every fixed action is pointwise bounded by the statewise maximum, giving Eq.~\ref{eq:information}. In the two-state case, a blind rule choosing $a_0$ with probability $r$ loses $p(1-r)\delta_0+(1-p)r\delta_1$ relative to the state-conditioned rule. Minimizing this affine expression over $r\in[0,1]$ gives Eq.~\ref{eq:blindness_gap}.
\end{proof}

This gap comes from a change in which computation is worth requesting, even with perfect knowledge of task quality. For a final agent step with expected correctness $Q(a)$ and expected remaining latency $D_z(a)$, action $a_1$ is preferred to $a_0$ precisely when
\begin{equation}
 D_z(a_0)-D_z(a_1)>\frac{\beta}{\lambda}\bigl(Q(a_0)-Q(a_1)\bigr),\qquad\lambda>0.
 \label{eq:switching}
\end{equation}
Here quality is independent of load: queueing alone can reverse the ranking. A model--reasoning pair can therefore become preferable without any change to the task. This decision-level result motivates refreshing the executor's state as the workflow progresses; \sys{} learns these choices from outcomes rather than requiring known $Q$ or $D$ functions. It characterizes optimal decisions under the stated assumptions, not a performance or deadline guarantee for the learned policy.

\section{Method}
\label{sec:method}

\begin{figure}[t]
  \centering
  \includegraphics[width=\textwidth]{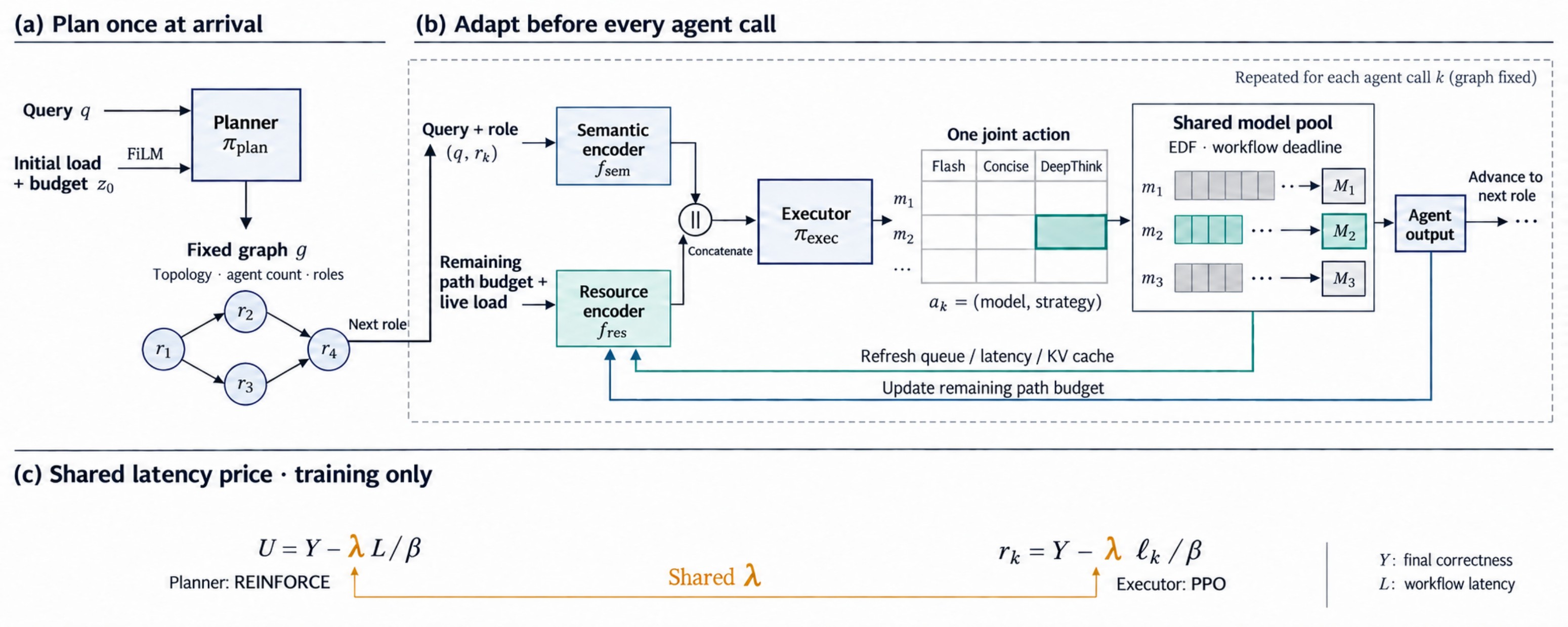}
  \caption{\sys{} couples planning with execution-time recourse. \textbf{(a)}~The query, initial load and budget determine a role graph fixed at arrival. \textbf{(b)}~Semantic and resource features feed one joint model--strategy decision; execution refreshes telemetry and remaining path budget. \textbf{(c)}~Planner utility and executor rewards share the latency price $\lambda$ during training.}
  \label{fig:methodology}
\end{figure}

\sys{} implements Eq.~\ref{eq:hierarchy} as \emph{planning with execution-time recourse}: the graph is fixed at arrival, but model--strategy choices can change before each call. Figure~\ref{fig:methodology} shows the two decision timescales, execution feedback and shared training penalty; Algorithm~\ref{alg:inframind} specifies the rollout and learning loop.

\begin{algorithm}[tbp]
  \caption{\sys{}: hierarchical allocation of reasoning}
  \label{alg:inframind}
  \small
  \begin{algorithmic}[1]
    \Require Model pool $\mathcal M$, strategies $\mathcal S$, policies $\pi_{\rm plan},\pi_{\rm exec}$, multiplier $\lambda$
    \For{each training batch}
      \State Collect concurrent query rollouts under sampled arrival rates and budgets:
      \For{each query $q_i$ with budget $\beta_i$ and arrival time $t_i$}
        \State Observe $\mathbf z_0$; sample graph $g_i\sim\pi_{\rm plan}(\cdot\mid\mathbf e_{q_i},\mathbf z_0)$
        \For{each agent step $k$ in the graph's execution order}
          \State Refresh $s_{ik}$ from the role, remaining path budget and serving telemetry
          \State Sample $(m_{ik},\sigma_{ik})\sim\pi_{\rm exec}(\cdot\mid s_{ik})$
          \State Execute on $m_{ik}$ with prompt strategy $\sigma_{ik}$ and EDF deadline $t_i+\beta_i$
          \State Record the state, action, probability and latency; update dependency-path costs
        \EndFor
      \State Aggregate under the same deadline; record correctness $Y_i$ and workflow latency $L_i$
      \EndFor
      \State Form $U_i=Y_i-\lambda L_i/\beta_i$ and $r_{ik}=Y_i-\lambda\ell_{ik}/\beta_i$
      \State Update planner by Eq.~\ref{eq:reinforce} and executor by Eq.~\ref{eq:ppo}
      \State Update the shared multiplier using workflow costs (Eq.~\ref{eq:dual})
    \EndFor
  \end{algorithmic}
\end{algorithm}
At inference, the same planning and execution loop runs with fixed policies; rewards and parameter updates are used only during training.

\subsection{Infrastructure-Aware Planner}
\label{sec:planner}

The planner controls the amount and structure of work admitted by each query. We retain MasRouter's cascaded controller (task classifier $\rightarrow$ collaboration $\rightarrow$ agent count $\rightarrow$ roles)~\citep{yue2025masrouter} and condition every head on $\mathbf{z}_0$ via Feature-wise Linear Modulation:
\begin{equation}
  \tilde{\mathbf{e}}_q = \boldsymbol{\gamma}(\mathbf{z}_0) \odot \mathbf{e}_q + \boldsymbol{\beta}(\mathbf{z}_0)
  \label{eq:film}
\end{equation}
An MLP maps $\mathbf{z}_0$ to $(\boldsymbol{\gamma},\boldsymbol{\beta})$, modulating the query embedding before the cascade. All four heads share the initial serving state, adapting graph structure to available capacity at arrival.

\subsection{Infrastructure-Aware Executor}
\label{sec:executor}

The executor makes the central fine-grained decision: \emph{which model--reasoning pair is worth its time cost now?} Model and prompt jointly determine the requested computation, while live load and remaining budget determine its affordability. Selecting them together allows time saved in one model's queue to be spent on deeper reasoning elsewhere. The policy combines a \emph{semantic pathway} (query and role) with a \emph{resource pathway} (remaining budget and live metrics); the LLM weights remain fixed.

The semantic and resource pathways produce 128- and 64-dimensional features, respectively, which are concatenated and projected into a shared representation $\mathbf h$. A single joint head gives $\pi_{\rm exec}(a_k\mid s_k)=\operatorname{softmax}(\mathbf W_{\rm act}\mathbf h)$ over all $N|\mathcal S|$ model--strategy pairs, alongside a value head $V(s_k)$. This parameterization can change the preferred reasoning strategy when either the model or its load changes. Strategies build on CoT prompting~\citep{wei2022chain}: \textbf{Flash} requests a direct answer, \textbf{Concise} requests 2--3 steps, and \textbf{DeepThink} requests thorough reasoning and verification. The policy learns when deeper reasoning~\citep{wang2023selfconsistency, yao2023tree, snell2025scaling} is affordable; architecture details and templates are in Appendix~\ref{app:repro}.

\subsection{Budget-Aware Priority Scheduling}
\label{sec:scheduler}

A tight-budget request can stall behind a relaxed one inside a model's queue. We attach an absolute deadline $t_{\text{arrive}}+\beta$ to each query; every agent call inherits it, and the model's waiting queue follows Earliest-Deadline-First (EDF)~\citep{liu1973scheduling}. Routing balances work across models, while EDF orders requests within each model.

EDF is an admission-ordering rule: calls already generating run to completion, and in-batch execution remains the serving layer's responsibility~\citep{yu2022orca, agrawal2024taming, sun2024llumnix}. Its role is to propagate workflow urgency to pending calls; the learned policies control their model and reasoning strategy.

\subsection{Joint Hierarchical RL Training}
\label{sec:training}

The planner and executor decide at different timescales and receive different learning signals, but share an outcome. With a fixed queue policy (EDF by default), we train the executor using the PPO clipped surrogate~\citep{schulman2017proximal} and the planner using baseline-normalized REINFORCE~\citep{williams1992simple}. One multiplier $\lambda$ prices latency for both:
\begin{equation}
  \mathcal{L}_{\text{exec}} = -\mathbb{E}\!\left[\min\!\left(\rho_k \hat{A}_k,\; \text{clip}(\rho_k, 1{-}\epsilon, 1{+}\epsilon)\hat{A}_k\right)\right] + 0.5\,\mathcal{L}_{\text{value}} - \alpha_H \mathcal{H}[\pi],
  \label{eq:ppo}
\end{equation}
\begin{equation}
  \mathcal{L}_{\text{plan}} = -\log \pi_{\text{plan}} \cdot \frac{U_i - \bar{U}}{\sigma_U} + \mathcal{L}_{\text{task}} + \alpha_{\text{VAE}} \mathcal{L}_{\text{VAE}},
  \label{eq:reinforce}
\end{equation}
Here $\rho_k$ is the new-to-rollout action probability ratio and $\hat A_k$ is the normalized advantage $r_k-V(s_k)$. The executor reward $r_k=Y-\lambda\ell_k/\beta$ credits the final answer to each step; planner utility $U_i=Y-\lambda L/\beta$ scores the complete workflow. The planner retains its task-classification loss and VAE regularizer. After each batch, $\lambda$ adapts to the mean normalized workflow cost $\bar C$ in the executor records:
\begin{equation}
  \lambda \leftarrow \text{clip}\!\left(\lambda + \eta_\lambda (\bar{C} - 1),\; 0,\; \lambda_{\max}\right).
  \label{eq:dual}
\end{equation}
This shared update couples the two timescales: expensive workflows increase the penalty on both creating work and executing it, while slack permits further investment in quality. Training sweeps budget tiers and arrival rates with inter-sweep queue draining (Appendix~\ref{app:hyperparams}).

\section{Experiments}
\label{sec:experiments}

\subsection{Setup}
\label{sec:setup}

\paragraph{Benchmarks and model pool.} We evaluate on five established benchmarks spanning code generation, mathematical reasoning and knowledge-intensive QA: MBPP~\citep{austin2021program}, HumanEval~\citep{chen2021evaluating}, GSM-Hard~\citep{gao2023pal}, MATH~\citep{hendrycks2021measuring} and MMLU-Pro~\citep{wang2024mmlu}, with splits in Appendix~\ref{app:datasets}. The pool is deliberately heterogeneous, a reasoning specialist, a code specialist, a mid-size generalist and two small generalists spanning a $10\times$ parameter range, so the executor faces a genuine capability-versus-capacity choice at every step: DeepSeek-R1-Distill-Qwen-32B~\citep{guo2025deepseek}, Mistral-Small-24B~\citep{mistral2025small}, Qwen2.5-Coder-14B~\citep{hui2024qwen2}, Llama-3.1-8B~\citep{grattafiori2024llama} and Llama-3.2-3B~\citep{meta2024llama32}, served via vLLM on two NVIDIA B200 GPUs.

\paragraph{Baselines.} We compare MoA~\citep{wang2406mixture} (ensemble), GPTSwarm~\citep{zhuge2024gptswarm} (learned graph, fixed at test time), and MasRouter~\citep{yue2025masrouter} (task-adaptive routing). Section~\ref{sec:robustness} adds paper-based reimplementations of the dynamic-workflow methods DyTopo~\citep{lu2026dytopo} and EvoMAS~\citep{xu2026evomas}.

\paragraph{Evaluation protocol.} All methods run on the shared pool under Poisson arrivals at $\{10,50,100\}$ requests/min. Table~\ref{tab:main_results} reports accuracy, mean latency and correct answers delivered within 300\,s. MoA, GPTSwarm and MasRouter use FCFS; \sys{} uses FCFS on MBPP and GSM-Hard, and EDF on MATH, HumanEval and MMLU-Pro. The common-EDF comparison separately controls scheduling with per-item deadline tiers. All methods share vLLM servers, the client and a 2{,}048-token output limit. We score final workflow answers by exact match on MATH, GSM-Hard and MMLU-Pro, and unit tests on HumanEval and MBPP. Appendices~\ref{app:datasets}--\ref{app:repro} specify counts, policy budgets, serving settings and prompts.

\subsection{Main Results}
\label{sec:results}

Our central outcome is \emph{delivered accuracy}: $N^{-1}\sum_i\mathbf{1}\{Y_i=1,\,L_i\leq\tau_i\}$, the fraction of evaluated workflows both correct and completed by the evaluation deadline $\tau_i$. We compute it from joint request-level records. Accuracy measures correctness independently of latency; SLO compliance measures timely completion. Table~\ref{tab:main_results} reports accuracy, latency and delivered accuracy; Appendix~\ref{app:stats} gives paired uncertainty for the larger-pool study.

\paragraph{The advantage persists with the same scheduler.}
On 500 MATH questions at 50 requests/min, every method uses EDF and the same per-item deadlines from $\{10,30,50,100,200,300\}$\,s. \sys{} delivers $33.8\%$ correctly and on time, versus $20.4\%$ for MasRouter and $0.2\%$ each for GPTSwarm and MoA: a $13.4$\,pp gain over the strongest baseline, with $2.1\times$ lower latency (160 vs.\ 329\,s). The advantage persists after matching queue policy (Appendix~\ref{app:extra}).

\begin{table}[t]
\centering
\caption{Accuracy (\%), mean latency (s) and delivered accuracy (Del, \% correct and completed within 300\,s). Correctness is pass@1. Each cell uses paired outcomes from the same evaluation records; counts and policy budgets are in Appendix~\ref{app:datasets}. \textbf{Bold} = best per column.}
\label{tab:main_results}
\resizebox{\textwidth}{!}{%
\begin{tabular}{l ccc ccc ccc ccc ccc}
\toprule
& \multicolumn{3}{c}{\textbf{MATH}} & \multicolumn{3}{c}{\textbf{MBPP}} & \multicolumn{3}{c}{\textbf{GSM-Hard}} & \multicolumn{3}{c}{\textbf{HumanEval}} & \multicolumn{3}{c}{\textbf{MMLU-Pro}} \\
\cmidrule(lr){2-4} \cmidrule(lr){5-7} \cmidrule(lr){8-10} \cmidrule(lr){11-13} \cmidrule(lr){14-16}
\textbf{Method} & Acc & Lat & Del & Acc & Lat & Del & Acc & Lat & Del & Acc & Lat & Del & Acc & Lat & Del \\
\midrule
\multicolumn{16}{c}{\textit{Low Load ($\mu = 10$ req/min)}} \\
\midrule
\sys{} (Ours) & \textbf{82.0} & \textbf{44} & \textbf{82.0} & \textbf{83.6} & \textbf{39} & \textbf{82.8} & \textbf{62.0} & \textbf{37} & \textbf{62.0} & \textbf{100.0} & \textbf{5} & \textbf{100.0} & \textbf{61.0} & 44 & \textbf{61.0} \\
MasRouter & 67.4 & 50 & 67.4 & 82.4 & 253 & 51.2 & 54.6 & 38 & 54.6 & 93.1 & 70 & 92.4 & 58.5 & 83 & 57.7 \\
GPTSwarm & 69.6 & 84 & 69.6 & 76.4 & 73 & 76.4 & 54.0 & 52 & 54.0 & \textbf{100.0} & 42 & \textbf{100.0} & 57.8 & \textbf{36} & 57.8 \\
MoA & 74.4 & 122 & 74.4 & 77.2 & 68 & 77.2 & 49.3 & 38 & 49.3 & 99.2 & 49 & 99.2 & 41.6 & 124 & 34.4 \\
\midrule
\multicolumn{16}{c}{\textit{Mid Load ($\mu = 50$ req/min)}} \\
\midrule
\sys{} (Ours) & \textbf{75.6} & \textbf{255} & \textbf{42.6} & 80.6 & \textbf{101} & \textbf{80.6} & \textbf{60.0} & \textbf{76} & \textbf{59.2} & \textbf{100.0} & \textbf{8} & \textbf{100.0} & \textbf{60.0} & \textbf{319} & \textbf{33.2} \\
MasRouter & 72.8 & 320 & 37.0 & \textbf{81.7} & 926 & 20.6 & 55.0 & 521 & 15.4 & 93.1 & 195 & 72.5 & 58.9 & 821 & 8.1 \\
GPTSwarm & 71.0 & 1209 & 4.6 & 76.6 & 1307 & 5.0 & 57.0 & 707 & 10.2 & \textbf{100.0} & 134 & \textbf{100.0} & 55.4 & 563 & 12.2 \\
MoA & 73.8 & 1145 & 2.2 & 75.2 & 969 & 4.0 & 49.1 & 474 & 8.4 & \textbf{100.0} & 155 & \textbf{100.0} & 46.2 & 1240 & 0.4 \\
\midrule
\multicolumn{16}{c}{\textit{High Load ($\mu = 100$ req/min)}} \\
\midrule
\sys{} (Ours) & \textbf{75.2} & \textbf{290} & \textbf{45.0} & 80.4 & \textbf{167} & \textbf{76.0} & \textbf{57.2} & \textbf{80} & \textbf{56.4} & \textbf{100.0} & \textbf{33} & \textbf{100.0} & \textbf{59.8} & \textbf{372} & \textbf{30.8} \\
MasRouter & 70.6 & 399 & 33.4 & \textbf{80.8} & 985 & 21.3 & 56.4 & 652 & 11.8 & 89.3 & 205 & 64.1 & 57.0 & 908 & 6.1 \\
GPTSwarm & 68.4 & 1325 & 2.0 & 78.6 & 1369 & 2.4 & 54.0 & 806 & 5.8 & \textbf{100.0} & 153 & \textbf{100.0} & 57.4 & 753 & 8.6 \\
MoA & 74.8 & 1160 & 0.4 & 75.8 & 993 & 1.6 & 50.0 & 524 & 6.0 & \textbf{100.0} & 177 & 95.4 & 47.8 & 1171 & 0.4 \\
\bottomrule
\end{tabular}}
\end{table}

\paragraph{Capacity becomes correct, timely answers.}
At 100 requests/min, delivered accuracy is $45.0\%$ versus the strongest baseline's $33.4\%$ on MATH, $76.0\%$ versus $21.3\%$ on MBPP, $56.4\%$ versus $11.8\%$ on GSM-Hard, and $30.8\%$ versus $8.6\%$ on MMLU-Pro; HumanEval ties at $100\%$. These gains combine correctness and timeliness. GSM-Hard deadline compliance is $98.4\%$ versus at most $20.6\%$, and MBPP compliance is $91.4\%$ versus at most $25.5\%$. At moderate and high load, \sys{} has the lowest mean latency on all five benchmarks.

\paragraph{Available time becomes answer quality.}
At low load, \sys{} has best or tied-best accuracy on all five benchmarks, exceeding the strongest baseline by $7.6$\,pp on MATH and $7.4$\,pp on GSM-Hard. On MATH, latency also falls from 122 to 44\,s relative to the most accurate baseline. At high load, it leads or ties on four benchmarks; on MBPP, accuracy is $80.4\%$ versus MasRouter's $80.8\%$, with $5.9\times$ lower latency.

\subsection{Analysis and Ablations}
\label{sec:analysis}

\paragraph{Joint model--reasoning selection spends time productively.}
Holding routing fixed and changing only the reasoning prompt on MMLU-Pro, DeepThink reaches $58.1\%$ accuracy versus Flash's $51.7\%$ ($1{,}200$ evaluations per strategy). This $6.4$-point difference establishes reasoning depth as a quality lever. The executor allocates that effort where capacity is available: its DeepThink calls select models with mean queue depth $11.0$, versus $31.9$ for Flash.

As load rises from 10 to 100 requests/min, the mean selected-model size falls from $22.7$B to $20.1$B, while DeepThink use rises from $36.4\%$ to $41.9\%$. This behavior is consistent with moving work to smaller available models and using deeper reasoning to offset the quality cost of smaller models; accuracy stays near $60\%$. The prompt comparison establishes the quality lever, while the following ablations distinguish learned selection from the contribution of serving signals.

\paragraph{Ablations separate selection from affordability.}
With planner and scheduler intact, learned model--strategy selection delivers $6.4$\,pp more MATH accuracy than random selection ($71.4\%$ vs.\ $65.0\%$). Live telemetry supplies the complementary affordability signal: on MMLU-Pro at 100 requests/min, the full executor delivers $28.0\%$ correct, on-time answers versus $21.8\%$ after removing its infrastructure inputs and retraining, at roughly half the latency (373 vs.\ 739\,s). Raw accuracy changes from $61.0\%$ without telemetry to $57.8\%$ with it: the policy trades some unconstrained accuracy for more answers delivered within the deadline (Appendix~\ref{app:ablation}). These ablations support learned selection and serving observations; load-conditioned planning, the shared penalty and joint rather than separate model--strategy selection remain design choices evaluated together.

\begin{figure}[tbp]
  \centering
  \includegraphics[width=0.36\textwidth]{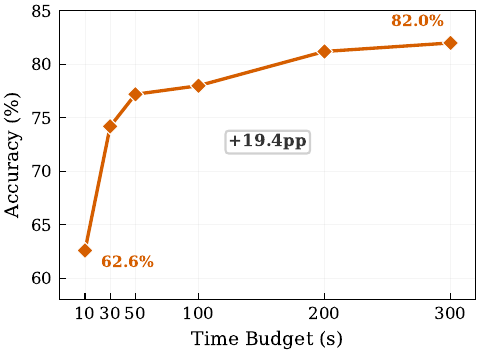}
  \caption{Accuracy vs.\ time budget on MATH ($\mu{=}10$ r/m). Accuracy rises from $62.6\%$ to $82.0\%$ ($+19.4$\,pp) as the executor shifts to larger models and DeepThink, emergent from end-to-end RL with no hand-coded rules.}
  \label{fig:budget_scaling}
\end{figure}

\paragraph{Adapting to the time budget.}
\label{sec:budget}
On MATH at low load, accuracy rises from $62.6\%$ to $82.0\%$ as the time budget increases (Figure~\ref{fig:budget_scaling}). The policy shifts toward larger models and DeepThink without budget-specific rules, showing that it uses available time to improve answer quality.

\subsection{Dynamic Workflows, Larger Pools and Arrival Replay}
\label{sec:robustness}

\paragraph{Orchestrators that adapt their structure at run time.}
Table~\ref{tab:newbaselines}a compares paper-based reimplementations of DyTopo~\citep{lu2026dytopo} and EvoMAS~\citep{xu2026evomas} on the same MATH problems, arrivals, budgets and EDF scheduler. \sys{} achieves $71.4\%$ accuracy at 160\,s, versus DyTopo's $69.0\%$ at 1529\,s and EvoMAS's $47.4\%$ at 519\,s. Their task-driven workflow adaptation complements the decision studied here: adapting requested computation to serving load.

\begin{table}[t]
  \centering
  \small
  \caption{MATH under dynamic workflows, larger pools and replayed Azure arrivals ($n{=}500$). DyTopo and EvoMAS are paper-based reimplementations. All methods within a setting share hardware, arrivals and budgets. ``Delivered'' is correct and on time as a percentage of all requests. Deadlines are per-item tiers for (a) and the 15-model study, and 60\,s for the bursty trace (Appendix~\ref{app:extra}).}
  \label{tab:newbaselines}
  \setlength{\tabcolsep}{4pt}
  \renewcommand{\arraystretch}{1.4}
  \begin{tabular}{@{}p{0.28\textwidth} l cccc@{}}
    \toprule
    \textbf{Setting} & \textbf{Method} & \textbf{Acc (\%)} & \textbf{Lat (s)} & \textbf{On time (\%)} & \textbf{Delivered (\%)} \\
    \midrule
    \multirow{3}{=}{\raggedright\textit{(a) Dynamic-topology orchestrators}\\5-model pool,\\Poisson $\mu{=}50$}
    & EvoMAS    & 47.4 & 519 & 2.2 & 1.6 \\
    & DyTopo    & 69.0 & 1529 & 0.4 & 0.0 \\
    & \sys{}    & \textbf{71.4} & \textbf{160} & \textbf{49.6} & \textbf{33.8} \\
    \midrule
    \textit{(b) Scale and arrival replay} & & & & & \\
    \multirow{2}{=}{\raggedright 15-model pool, Azure trace}
      & MASRouter & 70.2 & 188 & 14.8 & 11.6 \\
      & \sys{}    & \textbf{74.4} & \textbf{56} & \textbf{73.8} & \textbf{54.8} \\
    \cmidrule(l){2-6}
    \multirow{2}{=}{\raggedright 5-model pool, bursty Azure trace}
      & MASRouter & 73.2 & 232 & 18.6 & 13.6 \\
      & \sys{}    & \textbf{75.2} & \textbf{132} & \textbf{34.6} & \textbf{24.4} \\
    \bottomrule
  \end{tabular}
\end{table}

\paragraph{A larger model pool.}
We retrained both methods on \textbf{15 models across three B200s} and evaluated them on the same MATH problems with Azure arrival timings~\citep{patel2024splitwise} replayed at 50 requests/min, per-item deadlines in $\{10,30,50,100,200,300\}$\,s and identical hardware and EDF scheduling (Table~\ref{tab:newbaselines}b). \sys{} raises delivered accuracy from $11.6\%$ to $54.8\%$, a $4.7\times$ gain, while reducing mean latency from 188 to 56\,s. Paired McNemar tests favour it on accuracy ($p{=}0.047$) and on timely, correct delivery ($p{<}10^{-4}$; Appendix~\ref{app:stats}). MasRouter concentrates over half its calls on two models while alternatives remain idle: a larger pool makes available capacity useful only when the orchestrator can find it.

\paragraph{Responding to bursty arrivals.}
Under the bursty Azure trace and a 60\,s budget, \sys{} delivers $24.4\%$ of requests correctly and on time versus MasRouter's $13.6\%$, a $1.8\times$ gain, with mean latency of 132 versus 232\,s (Table~\ref{tab:newbaselines}b). The corresponding near-Poisson trace yields deadline attainment of $44.8\%$ and $39.6\%$, respectively; under bursts the gap widens to $34.6\%$ versus $18.6\%$. A separate telemetry-delay sweep shows stable performance over 0--10\,s delays at the tested operating point (Appendix~\ref{app:staleness}).

\subsection{Generalization to Blackbox and Hybrid Pools}
\label{sec:blackbox}

When server internals are hidden, we replace telemetry with two client-side proxies: an exponential moving average of response latency and recent requests divided by the provider's RPM limit. The executor architecture is unchanged (Appendix~\ref{app:blackboxdetails}).

Table~\ref{tab:blackbox_hybrid} evaluates GSM-Hard on hybrid and pure-API pools. At 100 requests/min, hybrid SLO compliance is $54.4\%$ versus MasRouter's $11.6\%$, with accuracy of $61.4\%$ versus $60.4\%$. Pure-API compliance is $50.4\%$ versus $30.6\%$, with accuracy of $64.6\%$ versus $61.2\%$. Client-observed responsiveness and rate-limit pressure can therefore guide orchestration beyond self-hosted infrastructure.

\begin{table}[tbp]
  \centering
  \small
  \caption{Hybrid (3 whitebox + 2 OpenRouter API, budget $100$\,s) and pure-API (5 OpenRouter, budget $500$\,s) pools on GSM-Hard. Acc (\%), Lat (s), SLO (\%). \textbf{Bold} = best per metric per load per pool.}
  \label{tab:blackbox_hybrid}
  \resizebox{\textwidth}{!}{%
  \begin{tabular}{ll ccc ccc ccc}
    \toprule
    & & \multicolumn{3}{c}{$\mu = 10$ r/m} & \multicolumn{3}{c}{$\mu = 50$ r/m} & \multicolumn{3}{c}{$\mu = 100$ r/m} \\
    \cmidrule(lr){3-5} \cmidrule(lr){6-8} \cmidrule(lr){9-11}
    \textbf{Pool} & \textbf{Method} & Acc & Lat & SLO & Acc & Lat & SLO & Acc & Lat & SLO \\
    \midrule
    \multirow{3}{*}{Hybrid}
      & \sys{}     & 62.4          & \textbf{14}  & \textbf{100.0} & \textbf{61.6} & \textbf{43}  & \textbf{94.4} & \textbf{61.4} & \textbf{107} & \textbf{54.4} \\
      & MASRouter  & \textbf{62.8} & 28           & 99.8           & 61.4          & 352          & 17.4          & 60.4          & 441          & 11.6 \\
      & GPTSwarm   & 59.4          & 42           & \textbf{100.0}          & 59.8          & 209          & 19.6          & 58.8          & 412          & 9.4 \\
    \midrule
    \multirow{3}{*}{Pure-API}
      & \sys{}     & 63.0          & \textbf{3}   & \textbf{100.0} & 63.7          & \textbf{370} & \textbf{68.1} & \textbf{64.6} & \textbf{492} & \textbf{50.4} \\
      & MASRouter  & 62.8          & 114          & 96.2           & 63.6          & 714          & 40.6          & 61.2          & 813          & 30.6 \\
      & GPTSwarm   & \textbf{63.6} & 70           & 99.8           & \textbf{64.6} & 686          & 23.6          & 64.0          & 902          & 11.6 \\
    \bottomrule
  \end{tabular}}
\end{table}

\section{Conclusion}
\label{sec:conclusion}

\sys{} allocates reasoning where it is both useful and affordable. Load-conditioned planning chooses the work to create, joint model--strategy selection adapts its execution, and a shared latency penalty connects them. Across five benchmarks, we observe gains in accuracy and timeliness. Matched-scheduler and larger-pool comparisons directly demonstrate the practical outcome: more correct answers delivered before their deadlines.

\clearpage
\subsubsection*{Reproducibility statement}

Evaluations use live vLLM servers or hosted APIs under concurrent traffic. Appendices~\ref{app:hyperparams}--\ref{app:extra} document training, dataset splits, model placement, serving settings, prompts, scoring and workload protocols; Appendix~\ref{app:stats} specifies the paired statistical tests. Algorithm~\ref{alg:inframind} details the rollout and learning loop, and Appendix~\ref{app:repro} gives the executor architecture, budget bookkeeping and exact reasoning prompts, so every component of \sys{} can be reimplemented from the paper.

\subsubsection*{Ethics statement}

This work uses public benchmarks, open-weight models and hosted APIs, involves no human subjects, and collects no personal data. Arrival replay uses timing information from public Azure traces. More efficient orchestration can improve access to timely model responses; applications remain responsible for the safety and reliability of the underlying models and generated outputs.

\subsubsection*{AI use statement}

We used generative AI tools to aid and polish writing, including copy-editing and improving clarity, and to draft sections of the paper. We also used these tools for retrieval and discovery, including finding related work and checking citation details against source records. For research execution, we used AI coding assistance to implement and debug parts of the experimental harness and analysis scripts. The authors are responsible for the research decisions and for verifying all AI-assisted content, including the accuracy of citations, code, results and claims in the final paper.


\bibliographystyle{iclr2027_conference}
\bibliography{references}

\begin{thebibliography}{50}
\providecommand{\natexlab}[1]{#1}
\providecommand{\url}[1]{\texttt{#1}}
\expandafter\ifx\csname urlstyle\endcsname\relax
  \providecommand{\doi}[1]{doi: #1}\else
  \providecommand{\doi}{doi: \begingroup \urlstyle{rm}\Url}\fi

\bibitem[Agrawal et~al.(2024)Agrawal, Kedia, Panwar, Mohan, Kwatra, Gulavani,
  Tumanov, and Ramjee]{agrawal2024taming}
Amey Agrawal, Nitin Kedia, Ashish Panwar, Jayashree Mohan, Nipun Kwatra,
  Bhargav Gulavani, Alexey Tumanov, and Ramachandran Ramjee.
\newblock Taming {Throughput-Latency} tradeoff in {LLM} inference with
  {Sarathi-Serve}.
\newblock In \emph{18th USENIX Symposium on Operating Systems Design and
  Implementation (OSDI 24)}, pp.\  117--134, Santa Clara, CA, July 2024. USENIX
  Association.
\newblock ISBN 978-1-939133-40-3.
\newblock URL
  \url{https://www.usenix.org/conference/osdi24/presentation/agrawal}.

\bibitem[Altman(1999)]{altman2021constrained}
Eitan Altman.
\newblock \emph{Constrained Markov Decision Processes}.
\newblock Stochastic Modeling. Chapman \& Hall/CRC, Boca Raton, 1999.
\newblock ISBN 9780849303821.
\newblock URL \url{https://worldcat.org/title/41258998}.

\bibitem[Austin et~al.(2021)Austin, Odena, Nye, Bosma, Michalewski, Dohan,
  Jiang, Cai, Terry, Le, et~al.]{austin2021program}
Jacob Austin, Augustus Odena, Maxwell Nye, Maarten Bosma, Henryk Michalewski,
  David Dohan, Ellen Jiang, Carrie Cai, Michael Terry, Quoc Le, et~al.
\newblock Program synthesis with large language models.
\newblock \emph{arXiv preprint arXiv:2108.07732}, 2021.

\bibitem[Chen et~al.(2025)Chen, Shi, Chen, and Guo]{chen2025kairos}
Jinyuan Chen, Jiuchen Shi, Quan Chen, and Minyi Guo.
\newblock Kairos: Low-latency multi-agent serving with shared llms and
  excessive loads in the public cloud.
\newblock \emph{arXiv preprint arXiv:2508.06948}, 2025.

\bibitem[Chen et~al.(2021)Chen, Tworek, Jun, Yuan, Pinto, Kaplan, Edwards,
  Burda, Joseph, Brockman, et~al.]{chen2021evaluating}
Mark Chen, Jerry Tworek, Heewoo Jun, Qiming Yuan, Henrique Ponde De~Oliveira
  Pinto, Jared Kaplan, Harri Edwards, Yuri Burda, Nicholas Joseph, Greg
  Brockman, et~al.
\newblock Evaluating large language models trained on code.
\newblock \emph{arXiv preprint arXiv:2107.03374}, 2021.

\bibitem[Gao et~al.(2023)Gao, Madaan, Zhou, Alon, Liu, Yang, Callan, and
  Neubig]{gao2023pal}
Luyu Gao, Aman Madaan, Shuyan Zhou, Uri Alon, Pengfei Liu, Yiming Yang, Jamie
  Callan, and Graham Neubig.
\newblock {PAL}: Program-aided language models.
\newblock In Andreas Krause, Emma Brunskill, Kyunghyun Cho, Barbara Engelhardt,
  Sivan Sabato, and Jonathan Scarlett (eds.), \emph{Proceedings of the 40th
  International Conference on Machine Learning}, volume 202 of
  \emph{Proceedings of Machine Learning Research}, pp.\  10764--10799. PMLR,
  23--29 Jul 2023.
\newblock URL \url{https://proceedings.mlr.press/v202/gao23f.html}.

\bibitem[Grattafiori et~al.(2024)Grattafiori, Dubey, Jauhri, Pandey, Kadian,
  Al-Dahle, Letman, Mathur, Schelten, Vaughan, et~al.]{grattafiori2024llama}
Aaron Grattafiori, Abhimanyu Dubey, Abhinav Jauhri, Abhinav Pandey, Abhishek
  Kadian, Ahmad Al-Dahle, Aiesha Letman, Akhil Mathur, Alan Schelten, Alex
  Vaughan, et~al.
\newblock The llama 3 herd of models.
\newblock \emph{arXiv preprint arXiv:2407.21783}, 2024.

\bibitem[Guo et~al.(2025)Guo, Yang, Zhang, Song, Wang, Zhu, Xu, Zhang, Ma, Bi,
  et~al.]{guo2025deepseek}
Daya Guo, Dejian Yang, Haowei Zhang, Junxiao Song, Peiyi Wang, Qihao Zhu,
  Runxin Xu, Ruoyu Zhang, Shirong Ma, Xiao Bi, et~al.
\newblock Deepseek-r1: Incentivizing reasoning capability in llms via
  reinforcement learning.
\newblock \emph{arXiv preprint arXiv:2501.12948}, 2025.

\bibitem[Hendrycks et~al.(2021)Hendrycks, Burns, Kadavath, Arora, Basart, Tang,
  Song, and Steinhardt]{hendrycks2021measuring}
Dan Hendrycks, Collin Burns, Saurav Kadavath, Akul Arora, Steven Basart, Eric
  Tang, Dawn Song, and Jacob Steinhardt.
\newblock Measuring mathematical problem solving with the math dataset.
\newblock In J.~Vanschoren and S.~Yeung (eds.), \emph{Proceedings of the Neural
  Information Processing Systems Track on Datasets and Benchmarks}, volume~1,
  2021.
\newblock URL
  \url{https://datasets-benchmarks-proceedings.neurips.cc/paper_files/paper/2021/file/be83ab3ecd0db773eb2dc1b0a17836a1-Paper-round2.pdf}.

\bibitem[Hong et~al.(2024)Hong, Zhuge, Chen, Zheng, Cheng, Wang, Zhang, wang,
  Yau, Lin, Zhou, Ran, Xiao, Wu, and Schmidhuber]{hong2023metagpt}
Sirui Hong, Mingchen Zhuge, Jonathan Chen, Xiawu Zheng, Yuheng Cheng, Jinlin
  Wang, Ceyao Zhang, zili wang, Steven Yau, Zijuan Lin, Liyang Zhou, Chenyu
  Ran, Lingfeng Xiao, Chenglin Wu, and J\"{u}rgen Schmidhuber.
\newblock Metagpt: Meta programming for a multi-agent collaborative framework.
\newblock In B.~Kim, Y.~Yue, S.~Chaudhuri, K.~Fragkiadaki, M.~Khan, and Y.~Sun
  (eds.), \emph{International Conference on Learning Representations}, volume
  2024, pp.\  23247--23275, 2024.
\newblock URL
  \url{https://proceedings.iclr.cc/paper_files/paper/2024/file/6507b115562bb0a305f1958ccc87355a-Paper-Conference.pdf}.

\bibitem[Hu et~al.(2025)Hu, Lu, and Clune]{hu2025automated}
Shengran Hu, Cong Lu, and Jeff Clune.
\newblock Automated design of agentic systems.
\newblock In Y.~Yue, A.~Garg, N.~Peng, F.~Sha, and R.~Yu (eds.),
  \emph{International Conference on Learning Representations}, volume 2025,
  pp.\  21344--21377, 2025.
\newblock URL
  \url{https://proceedings.iclr.cc/paper_files/paper/2025/file/36b7acf6f6010652b3f2a433774a66fe-Paper-Conference.pdf}.

\bibitem[Hui et~al.(2024)Hui, Yang, Cui, Yang, Liu, Zhang, Liu, Zhang, Yu, Lu,
  et~al.]{hui2024qwen2}
Binyuan Hui, Jian Yang, Zeyu Cui, Jiaxi Yang, Dayiheng Liu, Lei Zhang, Tianyu
  Liu, Jiajun Zhang, Bowen Yu, Keming Lu, et~al.
\newblock Qwen2. 5-coder technical report.
\newblock \emph{arXiv preprint arXiv:2409.12186}, 2024.

\bibitem[Kwon et~al.(2023)Kwon, Li, Zhuang, Sheng, Zheng, Yu, Gonzalez, Zhang,
  and Stoica]{kwon2023efficient}
Woosuk Kwon, Zhuohan Li, Siyuan Zhuang, Ying Sheng, Lianmin Zheng, Cody~Hao Yu,
  Joseph Gonzalez, Hao Zhang, and Ion Stoica.
\newblock Efficient memory management for large language model serving with
  pagedattention.
\newblock In \emph{Proceedings of the 29th Symposium on Operating Systems
  Principles}, SOSP '23, pp.\  611–626, New York, NY, USA, 2023. Association
  for Computing Machinery.
\newblock ISBN 9798400702297.
\newblock \doi{10.1145/3600006.3613165}.
\newblock URL \url{https://doi.org/10.1145/3600006.3613165}.

\bibitem[Li et~al.(2023)Li, Hammoud, Itani, Khizbullin, and
  Ghanem]{li2023camel}
Guohao Li, Hasan Hammoud, Hani Itani, Dmitrii Khizbullin, and Bernard Ghanem.
\newblock Camel: Communicative agents for "mind" exploration of large language
  model society.
\newblock In A.~Oh, T.~Naumann, A.~Globerson, K.~Saenko, M.~Hardt, and
  S.~Levine (eds.), \emph{Advances in Neural Information Processing Systems},
  volume~36, pp.\  51991--52008. Curran Associates, Inc., 2023.
\newblock \doi{10.52202/075280-2264}.
\newblock URL
  \url{https://proceedings.neurips.cc/paper_files/paper/2023/file/a3621ee907def47c1b952ade25c67698-Paper-Conference.pdf}.

\bibitem[Lin et~al.(2024)Lin, Han, Zhang, Yang, Yang, Chen, and
  Qiu]{lin2024parrot}
Chaofan Lin, Zhenhua Han, Chengruidong Zhang, Yuqing Yang, Fan Yang, Chen Chen,
  and Lili Qiu.
\newblock Parrot: Efficient serving of {LLM-based} applications with semantic
  variable.
\newblock In \emph{18th USENIX Symposium on Operating Systems Design and
  Implementation (OSDI 24)}, pp.\  929--945, Santa Clara, CA, July 2024. USENIX
  Association.
\newblock ISBN 978-1-939133-40-3.
\newblock URL
  \url{https://www.usenix.org/conference/osdi24/presentation/lin-chaofan}.

\bibitem[Liu \& Layland(1973)Liu and Layland]{liu1973scheduling}
C.~L. Liu and James~W. Layland.
\newblock Scheduling algorithms for multiprogramming in a hard-real-time
  environment.
\newblock \emph{J. ACM}, 20\penalty0 (1):\penalty0 46–61, January 1973.
\newblock ISSN 0004-5411.
\newblock \doi{10.1145/321738.321743}.
\newblock URL \url{https://doi.org/10.1145/321738.321743}.

\bibitem[Liu et~al.(2024)Liu, Zhang, Li, Liu, and Yang]{liu2024dylan}
Zijun Liu, Yanzhe Zhang, Peng Li, Yang Liu, and Diyi Yang.
\newblock A dynamic {LLM}-powered agent network for task-oriented agent
  collaboration.
\newblock In \emph{First Conference on Language Modeling}, 2024.
\newblock URL \url{https://openreview.net/forum?id=XII0Wp1XA9}.

\bibitem[Lu et~al.(2026)Lu, Hu, Zhao, and Cao]{lu2026dytopo}
Yuxing Lu, Yucheng Hu, Xukai Zhao, and Jiuxin Cao.
\newblock Dytopo: Dynamic topology routing for multi-agent reasoning via
  semantic matching.
\newblock \emph{arXiv preprint arXiv:2602.06039}, 2026.

\bibitem[Luo et~al.(2026)Luo, Shi, Cai, Zhang, Wong, Wang, Wang, Huang, Chen,
  Gonzalez, and Stoica]{luo2026agentix}
Michael Luo, Xiaoxiang Shi, Colin Cai, Tianjun Zhang, Justin Wong, Yichuan
  Wang, Chi Wang, Yanping Huang, Zhifeng Chen, Joseph~E. Gonzalez, and Ion
  Stoica.
\newblock Agentix: An efficient serving engine for {LLM} agents as general
  programs.
\newblock In \emph{23rd USENIX Symposium on Networked Systems Design and
  Implementation (NSDI 26)}, pp.\  2443--2459, Renton, WA, May 2026. USENIX
  Association.
\newblock ISBN 978-1-939133-54-0.
\newblock URL \url{https://www.usenix.org/conference/nsdi26/presentation/luo}.

\bibitem[McNemar(1947)]{mcnemar1947note}
Quinn McNemar.
\newblock Note on the sampling error of the difference between correlated
  proportions or percentages.
\newblock \emph{Psychometrika}, 12\penalty0 (2):\penalty0 153--157, 1947.

\bibitem[{Meta AI}(2024)]{meta2024llama32}
{Meta AI}.
\newblock {Llama-3.2-3B-Instruct} model card, 2024.
\newblock URL \url{https://huggingface.co/meta-llama/Llama-3.2-3B-Instruct}.

\bibitem[{Mistral AI}(2025)]{mistral2025small}
{Mistral AI}.
\newblock {Mistral-Small-24B-Instruct-2501} model card, 2025.
\newblock URL
  \url{https://huggingface.co/mistralai/Mistral-Small-24B-Instruct-2501}.

\bibitem[Ong et~al.(2025)Ong, Almahairi, Wu, Chiang, Wu, Gonzalez, Kadous, and
  Stoica]{ong2024routellm}
Isaac Ong, Amjad Almahairi, Vincent Wu, Wei-Lin Chiang, Tianhao Wu, Joseph~E
  Gonzalez, Mohammed Kadous, and Ion Stoica.
\newblock Routellm: Learning to route llms from preference data.
\newblock In \emph{International Conference on Learning Representations},
  volume 2025, pp.\  34433--34448, 2025.

\bibitem[Patel et~al.(2024)Patel, Choukse, Zhang, Shah, Goiri, Maleki, and
  Bianchini]{patel2024splitwise}
Pratyush Patel, Esha Choukse, Chaojie Zhang, Aashaka Shah, Íñigo Goiri, Saeed
  Maleki, and Ricardo Bianchini.
\newblock Splitwise: Efficient generative llm inference using phase splitting.
\newblock In \emph{ISCA}, June 2024.
\newblock URL
  \url{https://www.microsoft.com/en-us/research/publication/splitwise-efficient-generative-llm-inference-using-phase-splitting/}.

\bibitem[Qin et~al.(2025)Qin, Li, He, Cui, Ren, Zhang, Wu, Zheng, and
  Xu]{qin2025mooncake}
Ruoyu Qin, Zheming Li, Weiran He, Jialei Cui, Feng Ren, Mingxing Zhang, Yongwei
  Wu, Weimin Zheng, and Xinran Xu.
\newblock Mooncake: Trading more storage for less computation {\textemdash} a
  {KVCache-centric} architecture for serving {LLM} chatbot.
\newblock In \emph{23rd USENIX Conference on File and Storage Technologies
  (FAST 25)}, pp.\  155--170, Santa Clara, CA, February 2025. USENIX
  Association.
\newblock ISBN 978-1-939133-45-8.
\newblock URL \url{https://www.usenix.org/conference/fast25/presentation/qin}.

\bibitem[Reimers \& Gurevych(2019)Reimers and Gurevych]{reimers2019sentence}
Nils Reimers and Iryna Gurevych.
\newblock Sentence-{BERT}: Sentence embeddings using {S}iamese {BERT}-networks.
\newblock In Kentaro Inui, Jing Jiang, Vincent Ng, and Xiaojun Wan (eds.),
  \emph{Proceedings of the 2019 Conference on Empirical Methods in Natural
  Language Processing and the 9th International Joint Conference on Natural
  Language Processing (EMNLP-IJCNLP)}, pp.\  3982--3992, Hong Kong, China,
  November 2019. Association for Computational Linguistics.
\newblock \doi{10.18653/v1/D19-1410}.
\newblock URL \url{https://aclanthology.org/D19-1410/}.

\bibitem[Schulman et~al.(2017)Schulman, Wolski, Dhariwal, Radford, and
  Klimov]{schulman2017proximal}
John Schulman, Filip Wolski, Prafulla Dhariwal, Alec Radford, and Oleg Klimov.
\newblock Proximal policy optimization algorithms.
\newblock \emph{arXiv preprint arXiv:1707.06347}, 2017.

\bibitem[Snell et~al.(2025)Snell, Lee, Xu, and Kumar]{snell2025scaling}
Charlie Snell, Jaehoon Lee, Kelvin Xu, and Aviral Kumar.
\newblock Scaling llm test-time compute optimally can be more effective than
  scaling parameters for reasoning.
\newblock In Y.~Yue, A.~Garg, N.~Peng, F.~Sha, and R.~Yu (eds.),
  \emph{International Conference on Learning Representations}, volume 2025,
  pp.\  10131--10165, 2025.
\newblock URL
  \url{https://proceedings.iclr.cc/paper_files/paper/2025/file/1b623663fd9b874366f3ce019fdfdd44-Paper-Conference.pdf}.

\bibitem[Sun et~al.(2024)Sun, Huang, Zhao, Xiao, Zhang, Li, and
  Lin]{sun2024llumnix}
Biao Sun, Ziming Huang, Hanyu Zhao, Wencong Xiao, Xinyi Zhang, Yong Li, and Wei
  Lin.
\newblock Llumnix: Dynamic scheduling for large language model serving.
\newblock In \emph{18th USENIX Symposium on Operating Systems Design and
  Implementation (OSDI 24)}, pp.\  173--191, Santa Clara, CA, July 2024. USENIX
  Association.
\newblock ISBN 978-1-939133-40-3.
\newblock URL
  \url{https://www.usenix.org/conference/osdi24/presentation/sun-biao}.

\bibitem[Wang et~al.(2025{\natexlab{a}})Wang, Wang, Athiwaratkun, Zhang, and
  Zou]{wang2406mixture}
Junlin Wang, Jue Wang, Ben Athiwaratkun, Ce~Zhang, and James~Y Zou.
\newblock Mixture-of-agents enhances large language model capabilities.
\newblock In \emph{International Conference on Learning Representations},
  volume 2025, pp.\  33944--33963, 2025{\natexlab{a}}.

\bibitem[Wang et~al.(2023)Wang, Wei, Schuurmans, Le, Chi, Narang, Chowdhery,
  and Zhou]{wang2023selfconsistency}
Xuezhi Wang, Jason Wei, Dale Schuurmans, Quoc~V. Le, Ed~H. Chi, Sharan Narang,
  Aakanksha Chowdhery, and Denny Zhou.
\newblock Self-consistency improves chain of thought reasoning in language
  models.
\newblock In \emph{ICLR 2023}, 2023.
\newblock URL \url{https://arxiv.org/abs/2203.11171}.

\bibitem[Wang et~al.(2024)Wang, Ma, Zhang, Ni, Chandra, Guo, Ren, Arulraj, He,
  Jiang, Li, Ku, Wang, Zhuang, Fan, Yue, and Chen]{wang2024mmlu}
Yubo Wang, Xueguang Ma, Ge~Zhang, Yuansheng Ni, Abhranil Chandra, Shiguang Guo,
  Weiming Ren, Aaran Arulraj, Xuan He, Ziyan Jiang, Tianle Li, Max Ku, Kai
  Wang, Alex Zhuang, Rongqi Fan, Xiang Yue, and Wenhu Chen.
\newblock Mmlu-pro: A more robust and challenging multi-task language
  understanding benchmark.
\newblock In A.~Globerson, L.~Mackey, D.~Belgrave, A.~Fan, U.~Paquet,
  J.~Tomczak, and C.~Zhang (eds.), \emph{Advances in Neural Information
  Processing Systems}, volume~37, pp.\  95266--95290. Curran Associates, Inc.,
  2024.
\newblock \doi{10.52202/079017-3018}.
\newblock URL
  \url{https://proceedings.neurips.cc/paper_files/paper/2024/file/ad236edc564f3e3156e1b2feafb99a24-Paper-Datasets_and_Benchmarks_Track.pdf}.

\bibitem[Wang et~al.(2025{\natexlab{b}})Wang, Chen, Li, Kang, Fang, Zhou,
  Zheng, Tang, He, Guo, Wang, Wang, Zhou, and Chu]{wang2025burstgpt}
Yuxin Wang, Yuhan Chen, Zeyu Li, Xueze Kang, Yuchu Fang, Yeju Zhou, Yang Zheng,
  Zhenheng Tang, Xin He, Rui Guo, Xin Wang, Qiang Wang, Amelie~Chi Zhou, and
  Xiaowen Chu.
\newblock Burstgpt: A real-world workload dataset to optimize llm serving
  systems.
\newblock In \emph{Proceedings of the 31st ACM SIGKDD Conference on Knowledge
  Discovery and Data Mining V.2}, KDD '25, pp.\  5831–5841, New York, NY,
  USA, 2025{\natexlab{b}}. Association for Computing Machinery.
\newblock ISBN 9798400714542.
\newblock \doi{10.1145/3711896.3737413}.
\newblock URL \url{https://doi.org/10.1145/3711896.3737413}.

\bibitem[Wei et~al.(2022)Wei, Wang, Schuurmans, Bosma, ichter, Xia, Chi, Le,
  and Zhou]{wei2022chain}
Jason Wei, Xuezhi Wang, Dale Schuurmans, Maarten Bosma, brian ichter, Fei Xia,
  Ed~Chi, Quoc~V Le, and Denny Zhou.
\newblock Chain-of-thought prompting elicits reasoning in large language
  models.
\newblock In S.~Koyejo, S.~Mohamed, A.~Agarwal, D.~Belgrave, K.~Cho, and A.~Oh
  (eds.), \emph{Advances in Neural Information Processing Systems}, volume~35,
  pp.\  24824--24837. Curran Associates, Inc., 2022.
\newblock \doi{10.52202/068431-1800}.
\newblock URL
  \url{https://proceedings.neurips.cc/paper_files/paper/2022/file/9d5609613524ecf4f15af0f7b31abca4-Paper-Conference.pdf}.

\bibitem[Williams(1992)]{williams1992simple}
Ronald~J. Williams.
\newblock Simple statistical gradient-following algorithms for connectionist
  reinforcement learning.
\newblock \emph{Machine Learning}, 8\penalty0 (3):\penalty0 229--256, 1992.
\newblock \doi{10.1007/BF00992696}.
\newblock URL \url{https://doi.org/10.1007/BF00992696}.

\bibitem[Wilson(1927)]{wilson1927probable}
Edwin~B. Wilson.
\newblock Probable inference, the law of succession, and statistical inference.
\newblock \emph{Journal of the American Statistical Association}, 22\penalty0
  (158):\penalty0 209--212, 1927.
\newblock ISSN 01621459, 1537274X.
\newblock URL \url{http://www.jstor.org/stable/2276774}.

\bibitem[Wu et~al.(2024)Wu, Bansal, Zhang, Wu, Li, Zhu, Jiang, Zhang, Zhang,
  Awadallah, White, Burger, and Wang]{wu2024autogen}
Qingyun Wu, Gagan Bansal, Jieyu Zhang, Yiran Wu, Beibin Li, Erkang~(Eric) Zhu,
  Li~Jiang, Xiaoyun Zhang, Shaokun Zhang, Ahmed Awadallah, Ryen~W. White, Doug
  Burger, and Chi Wang.
\newblock Autogen: Enabling next-gen llm applications via multi-agent
  conversation.
\newblock In \emph{COLM 2024}, August 2024.
\newblock URL
  \url{https://www.microsoft.com/en-us/research/publication/autogen-enabling-next-gen-llm-applications-via-multi-agent-conversation-framework/}.

\bibitem[Xiang et~al.(2026)Xiang, Li, Qian, Zhang, Yu, Zhai, Jin, and
  Zhou]{xiang2026servegen}
Yuxing Xiang, Xue Li, Kun Qian, Yan Zhang, Wenyuan Yu, Ennan Zhai, Xin Jin, and
  Jingren Zhou.
\newblock {ServeGen}: Workload characterization and generation of large
  language model serving in production.
\newblock In \emph{23rd USENIX Symposium on Networked Systems Design and
  Implementation (NSDI 26)}, pp.\  1845--1859, Renton, WA, May 2026. USENIX
  Association.
\newblock ISBN 978-1-939133-54-0.
\newblock URL
  \url{https://www.usenix.org/conference/nsdi26/presentation/xiang-servegen}.

\bibitem[Xu et~al.(2026)Xu, Ke, Liu, Wei, Shao, Guo, and Yu]{xu2026evomas}
Chengdong Xu, Kaiqiang Ke, Ziheng Liu, Jiaqi Wei, Zibo Shao, Weile Guo, and
  Chao Yu.
\newblock Evomas: Learning execution-time workflows for multi-agent systems.
\newblock \emph{arXiv preprint arXiv:2605.08769}, 2026.

\bibitem[Xue et~al.(2026)Xue, Lou, Xing, and Huang]{xue2026r2}
Jiaqi Xue, Qian Lou, Jiarong Xing, and Heng Huang.
\newblock R2-router: A new paradigm for llm routing with reasoning.
\newblock \emph{arXiv preprint arXiv:2602.02823}, 2026.

\bibitem[Yao et~al.(2023)Yao, Yu, Zhao, Shafran, Griffiths, Cao, and
  Narasimhan]{yao2023tree}
Shunyu Yao, Dian Yu, Jeffrey Zhao, Izhak Shafran, Tom Griffiths, Yuan Cao, and
  Karthik Narasimhan.
\newblock Tree of thoughts: Deliberate problem solving with large language
  models.
\newblock In A.~Oh, T.~Naumann, A.~Globerson, K.~Saenko, M.~Hardt, and
  S.~Levine (eds.), \emph{Advances in Neural Information Processing Systems},
  volume~36, pp.\  11809--11822. Curran Associates, Inc., 2023.
\newblock \doi{10.52202/075280-0517}.
\newblock URL
  \url{https://proceedings.neurips.cc/paper_files/paper/2023/file/271db9922b8d1f4dd7aaef84ed5ac703-Paper-Conference.pdf}.

\bibitem[Yu et~al.(2022)Yu, Jeong, Kim, Kim, and Chun]{yu2022orca}
Gyeong-In Yu, Joo~Seong Jeong, Geon-Woo Kim, Soojeong Kim, and Byung-Gon Chun.
\newblock Orca: A distributed serving system for {Transformer-Based} generative
  models.
\newblock In \emph{16th USENIX Symposium on Operating Systems Design and
  Implementation (OSDI 22)}, pp.\  521--538, Carlsbad, CA, July 2022. USENIX
  Association.
\newblock ISBN 978-1-939133-28-1.
\newblock URL \url{https://www.usenix.org/conference/osdi22/presentation/yu}.

\bibitem[Yuan et~al.(2026)Yuan, Zuo, Wang, Chen, Tan, and Yu]{yuan2026dualmap}
Ying Yuan, Pengfei Zuo, Bo~Wang, Zhangyu Chen, Zhipeng Tan, and Zhou Yu.
\newblock Dualmap: Enabling both cache affinity and load balancing for
  distributed {LLM} serving.
\newblock In \emph{The Fourteenth International Conference on Learning
  Representations}, 2026.
\newblock URL \url{https://openreview.net/forum?id=zCadrJ32Xn}.

\bibitem[Yue et~al.(2025)Yue, Zhang, Liu, Wan, Wang, Cheng, and
  Qi]{yue2025masrouter}
Yanwei Yue, Guibin Zhang, Boyang Liu, Guancheng Wan, Kun Wang, Dawei Cheng, and
  Yiyan Qi.
\newblock {M}as{R}outer: Learning to route {LLM}s for multi-agent systems.
\newblock In Wanxiang Che, Joyce Nabende, Ekaterina Shutova, and Mohammad~Taher
  Pilehvar (eds.), \emph{Proceedings of the 63rd Annual Meeting of the
  Association for Computational Linguistics (Volume 1: Long Papers)}, pp.\
  15549--15572, Vienna, Austria, July 2025. Association for Computational
  Linguistics.
\newblock ISBN 979-8-89176-251-0.
\newblock \doi{10.18653/v1/2025.acl-long.757}.
\newblock URL \url{https://aclanthology.org/2025.acl-long.757/}.

\bibitem[Zhang et~al.(2025{\natexlab{a}})Zhang, Yue, Li, Yun, Wan, Wang, Cheng,
  Yu, and Chen]{zhang2025agentprune}
Guibin Zhang, Yanwei Yue, Zhixun Li, Sukwon Yun, Guancheng Wan, Kun Wang, Dawei
  Cheng, Jeffrey Yu, and Tianlong Chen.
\newblock Cut the crap: An economical communication pipeline for llm-based
  multi-agent systems.
\newblock In Y.~Yue, A.~Garg, N.~Peng, F.~Sha, and R.~Yu (eds.),
  \emph{International Conference on Learning Representations}, volume 2025,
  pp.\  75389--75428, 2025{\natexlab{a}}.
\newblock URL
  \url{https://proceedings.iclr.cc/paper_files/paper/2025/file/bbc461518c59a2a8d64e70e2c38c4a0e-Paper-Conference.pdf}.

\bibitem[Zhang et~al.(2025{\natexlab{b}})Zhang, Xiang, Yu, Teng, Chen, Chen,
  Zhuge, Cheng, Hong, Wang, Zheng, Liu, Luo, and Wu]{zhang2025aflow}
Jiayi Zhang, Jinyu Xiang, Zhaoyang Yu, Fengwei Teng, XiongHui Chen, Jiaqi Chen,
  Mingchen Zhuge, Xin Cheng, Sirui Hong, Jinlin Wang, Bingnan Zheng, Bang Liu,
  Yuyu Luo, and Chenglin Wu.
\newblock Aflow: Automating agentic workflow generation.
\newblock In Y.~Yue, A.~Garg, N.~Peng, F.~Sha, and R.~Yu (eds.),
  \emph{International Conference on Learning Representations}, volume 2025,
  pp.\  34040--34077, 2025{\natexlab{b}}.
\newblock URL
  \url{https://proceedings.iclr.cc/paper_files/paper/2025/file/5492ecbce4439401798dcd2c90be94cd-Paper-Conference.pdf}.

\bibitem[Zhang et~al.(2024)Zhang, Huang, Taga, Joe-Wong, Oymak, and
  Chen]{NEURIPS2024_a6deba3b}
Xuechen Zhang, Zijian Huang, Ege~Onur Taga, Carlee Joe-Wong, Samet Oymak, and
  Jiasi Chen.
\newblock Efficient contextual llm cascades through budget-constrained policy
  learning.
\newblock In A.~Globerson, L.~Mackey, D.~Belgrave, A.~Fan, U.~Paquet,
  J.~Tomczak, and C.~Zhang (eds.), \emph{Advances in Neural Information
  Processing Systems}, volume~37, pp.\  91691--91722. Curran Associates, Inc.,
  2024.
\newblock \doi{10.52202/079017-2910}.
\newblock URL
  \url{https://proceedings.neurips.cc/paper_files/paper/2024/file/a6deba3b2408af45b3f9994c2152b862-Paper-Conference.pdf}.

\bibitem[Zheng et~al.(2024)Zheng, Yin, Xie, Sun, Huang, Yu, Cao, Kozyrakis,
  Stoica, Gonzalez, Barrett, and Sheng]{zheng2024sglang}
Lianmin Zheng, Liangsheng Yin, Zhiqiang Xie, Chuyue Sun, Jeff Huang, Cody~Hao
  Yu, Shiyi Cao, Christos Kozyrakis, Ion Stoica, Joseph~E. Gonzalez, Clark
  Barrett, and Ying Sheng.
\newblock Sglang: Efficient execution of structured language model programs.
\newblock In A.~Globerson, L.~Mackey, D.~Belgrave, A.~Fan, U.~Paquet,
  J.~Tomczak, and C.~Zhang (eds.), \emph{Advances in Neural Information
  Processing Systems}, volume~37, pp.\  62557--62583. Curran Associates, Inc.,
  2024.
\newblock \doi{10.52202/079017-2000}.
\newblock URL
  \url{https://proceedings.neurips.cc/paper_files/paper/2024/file/724be4472168f31ba1c9ac630f15dec8-Paper-Conference.pdf}.

\bibitem[Zhong et~al.(2024)Zhong, Liu, Chen, Hu, Zhu, Liu, Jin, and
  Zhang]{zhong2024distserve}
Yinmin Zhong, Shengyu Liu, Junda Chen, Jianbo Hu, Yibo Zhu, Xuanzhe Liu, Xin
  Jin, and Hao Zhang.
\newblock {DistServe}: Disaggregating prefill and decoding for
  goodput-optimized large language model serving.
\newblock In \emph{18th USENIX Symposium on Operating Systems Design and
  Implementation (OSDI 24)}, pp.\  193--210, Santa Clara, CA, July 2024. USENIX
  Association.
\newblock ISBN 978-1-939133-40-3.
\newblock URL
  \url{https://www.usenix.org/conference/osdi24/presentation/zhong-yinmin}.

\bibitem[Zhuge et~al.(2024)Zhuge, Wang, Kirsch, Faccio, Khizbullin, and
  Schmidhuber]{zhuge2024gptswarm}
Mingchen Zhuge, Wenyi Wang, Louis Kirsch, Francesco Faccio, Dmitrii Khizbullin,
  and J\"{u}rgen Schmidhuber.
\newblock {GPTS}warm: Language agents as optimizable graphs.
\newblock In Ruslan Salakhutdinov, Zico Kolter, Katherine Heller, Adrian
  Weller, Nuria Oliver, Jonathan Scarlett, and Felix Berkenkamp (eds.),
  \emph{Proceedings of the 41st International Conference on Machine Learning},
  volume 235 of \emph{Proceedings of Machine Learning Research}, pp.\
  62743--62767. PMLR, 21--27 Jul 2024.
\newblock URL \url{https://proceedings.mlr.press/v235/zhuge24a.html}.

\end{thebibliography}

\clearpage
\appendix
\let\appendixsection\section
\renewcommand{\section}{\FloatBarrier\appendixsection}

\section{Training Hyperparameters}
\label{app:hyperparams}

\begin{table}[h]
  \centering
  \small
  \caption{Training hyperparameters for \sys{}.}
  \begin{tabular}{ll}
    \toprule
    \textbf{Parameter} & \textbf{Value} \\
    \midrule
    Optimizer & Adam, lr $3 \times 10^{-4}$ \\
    Batch size & 64 \\
    Max epochs & 40 (early stopping, patience 12) \\
    LR scheduler & ReduceLROnPlateau (patience 3, factor 0.5) \\
    \midrule
    PPO clip $\epsilon$ & 0.2 \\
    PPO mini-epochs $K$ & 3 \\
    Entropy coefficient $\alpha_H$ & 0.10 \\
    \midrule
    Initial $\lambda_0$ & 0.2 \\
    $\lambda$ learning rate $\eta_\lambda$ & 0.001 \\
    $\lambda_{\max}$ & 1.0 \\
    \midrule
    Budget tiers (s) & \{10, 30, 50, 100, 200, 300\} \\
    Arrival rates (req/min) & \{10, 30, 50, 100, 200\} \\
    vLLM \texttt{max\_num\_seqs} & 16 (all models) \\
    Total parameters & $\sim$471K \\
    \bottomrule
  \end{tabular}
\end{table}

\section{Dataset Details}
\label{app:datasets}

\begin{table}[h]
  \centering
  \small
  \caption{Dataset splits used in experiments. HumanEval has 164 problems in total; 33 are held out for training and 131 form the evaluation set.}
  \label{tab:datasets}
  \begin{tabular}{lccc}
    \toprule
    \textbf{Dataset} & \textbf{Train} & \textbf{Val} & \textbf{Test} \\
    \midrule
    MATH~\citep{hendrycks2021measuring} & 519 & 131 & 500 \\
    MBPP~\citep{austin2021program} & 374 & 94 & 500 \\
    GSM-Hard~\citep{gao2023pal} & 500 & 125 & 500 \\
    HumanEval~\citep{chen2021evaluating} & 33 & -- & 131 \\
    MMLU-Pro~\citep{wang2024mmlu} & 500 & 70 & 500 \\
    \bottomrule
  \end{tabular}
\end{table}

\paragraph{Main-table records and budgets.} Table~\ref{tab:main_results} scores timeliness against 300\,s for every method. \sys{}'s policy input budget is 300\,s on MATH, GSM-Hard and MMLU-Pro, and 100\,s on MBPP and HumanEval. The common-EDF and trace studies instead score against each request's assigned budget. Percentages use recorded workflow outcomes, not the nominal split size. At rates 10/50/100, counts are 500 per cell except: HumanEval, 131 throughout; \sys{} on MBPP, 499/500/499; MasRouter on MBPP, 500/496/494; MasRouter on MMLU-Pro, 496/496/493; and MoA on GSM-Hard, 499/499/500. The completed MBPP evaluation is used without pooling its earlier interrupted run. Missing records are not imputed; these counts do not represent independent training seeds.

\paragraph{GSM-Hard configurations.} The main-table GSM-Hard rows retain the existing load-specific configurations: the load-conditioned planner at 10 requests/min, and fixed latency penalties $\lambda=0.02$ and $0.04$ at 50 and 100 requests/min, respectively. All three use FCFS. These results evaluate configured policies; the paired telemetry and executor ablations provide the controlled mechanism comparisons.

\section{Reproducibility: Serving Environment, Prompts and Harness}
\label{app:repro}

\paragraph{Executor architecture.} The semantic pathway is $\mathbf h_{\rm sem}=\operatorname{LN}(\operatorname{ReLU}(\mathbf W_{\rm sem}[\mathbf e_q\|\mathbf e_{r_k}]))\in\mathbb R^{128}$. The resource pathway embeds budget and telemetry separately: $\mathbf h_{\rm bud}=\operatorname{ReLU}(\mathbf W_{\rm bud}b_k)$ and $\mathbf h_{\rm sys}=\operatorname{ReLU}(\mathbf W_{\rm sys}[\mathbf d^{\rm queue}\|\mathbf d^{\rm e2e}\|\mathbf d^{\rm kv}])$, each in $\mathbb R^{16}$. Their concatenation gives $\mathbf h_{\rm res}=\operatorname{LN}(\operatorname{ReLU}(\mathbf W_{\rm res}[\mathbf h_{\rm bud}\|\mathbf h_{\rm sys}]))\in\mathbb R^{64}$. The merged representation $\mathbf h=\operatorname{ReLU}(\mathbf W_{\rm merge}[\mathbf h_{\rm sem}\|\mathbf h_{\rm res}])$ feeds the joint action softmax and linear value head $V(s_k)=\mathbf w_{\rm val}^{\top}\mathbf h$.

\paragraph{Budget bookkeeping.} Before each agent call, the executor tracks the maximum cumulative latency along its predecessor paths, carrying the preceding round's critical-path cost into the next round. For path cost $d_k$, its budget feature is $b_k=\log(\max\{\beta-d_k,1\})/\log(300)$. Admission priority uses the absolute workflow deadline $t_{\rm arrive}+\beta$; the training cost and reported latency use measured elapsed workflow time $L$.

\paragraph{Placement and memory.} All five models remain resident on two B200s, each in a separate vLLM server with tensor-parallel size 1. Table~\ref{tab:placement} gives the fixed GPU memory allocation fractions; the remaining memory provides serving headroom. Models are neither swapped nor moved during evaluation.

\begin{table}[h]
  \centering
  \small
  \caption{Model placement and GPU memory fractions for the main pool. Every server uses tensor-parallel size 1.}
  \label{tab:placement}
  \begin{tabular}{llcc}
    \toprule
    \textbf{GPU} & \textbf{Model} & \textbf{Memory fraction} & \textbf{GPU total} \\
    \midrule
    \multirow{2}{*}{0} & DeepSeek-R1-Distill-Qwen-32B & 0.50 & \multirow{2}{*}{0.68} \\
                       & Qwen2.5-Coder-14B            & 0.18 & \\
    \midrule
    \multirow{3}{*}{1} & Mistral-Small-24B            & 0.30 & \multirow{3}{*}{0.57} \\
                       & Llama-3.1-8B                 & 0.15 & \\
                       & Llama-3.2-3B                 & 0.12 & \\
    \bottomrule
  \end{tabular}
\end{table}

\paragraph{Serving configuration.} Identical for every method and every arm: \texttt{bfloat16}, \texttt{-{}-max-model-len 16384}, \texttt{-{}-max-num-seqs 16}, \texttt{-{}-swap-space 16}, \texttt{-{}-no-enable-prefix-caching}, and vLLM's native \texttt{-{}-scheduling-policy priority}. The maximum output length is 2048 tokens for every model, every baseline and every reasoning strategy, so no method receives a larger generation allowance than another.

\paragraph{Queue implementation.} We add no queue of our own. The scheduler sets each request's \texttt{priority} field and vLLM reorders its own waiting queue; requests already running are never preempted or evicted. None of the evaluation is simulated.

\paragraph{Request lengths (MATH, measured).} \sys{} issues 3.6 LLM calls per request on average, totalling roughly 3.7k prompt and 2.5k completion tokens. MasRouter issues 5.3 calls, totalling roughly 3.9k prompt and 3.5k completion tokens. Per call, prompts average 735 tokens (p95 2357) and completions 670 tokens (p95 2023).

\paragraph{Reasoning templates.} The three strategies are short system prefixes, prepended to each call with the agent's role appended. No model is modified or fine-tuned, and the templates are identical across benchmarks:

\begin{itemize}\itemsep2pt
  \item \textbf{Flash:} \emph{``Respond in Flash mode. Go straight to the answer. Do not explain your reasoning or show intermediate steps. Output only the final answer in the required format.''}
  \item \textbf{Concise:} \emph{``Respond in Concise mode. Briefly outline your key reasoning in 2-3 short points, then state the final answer. Do not repeat the question or add unnecessary commentary.''}
  \item \textbf{DeepThink:} \emph{``Respond in DeepThink mode. Reason thoroughly and step-by-step. State your assumptions, show all intermediate work, verify your result, and provide a well-justified final answer.''}
\end{itemize}

The strategies differ only in what the prompt asks the model to produce, never in what the server permits it to produce: the token limit is 2048 output tokens within a 16384-token context for all three.

\paragraph{Evaluation harness.} We score one final workflow answer per problem; internal calls, reviews and aggregation follow each method's workflow. Thus pass@1 is a workflow solve rate, including internal voting where used. Graders use exact match for MATH, GSM-Hard and MMLU-Pro, and benchmark unit tests for HumanEval and MBPP. Main-pool methods share servers, client, token limits and a 3600\,s request timeout. HumanEval uses the 131-problem test split in Table~\ref{tab:datasets}; its saturated accuracy makes latency the distinguishing metric for the best-scoring methods.

\section{Additional Baselines, Scale, and Traces}
\label{app:extra}

\paragraph{Matched orchestration comparisons.} The additional MATH comparisons use 500 problems at 50 requests/min and budgets drawn from $\{10,30,50,100,200,300\}$\,s by item identifier, identical across methods. DyTopo and EvoMAS are paper-based reimplementations. All arms in the common-EDF comparison and Table~\ref{tab:newbaselines}a use the same pool, client and EDF queue policy; timeliness is scored against each request's assigned budget. This budget-conditioned run is separate from Table~\ref{tab:main_results}.

\paragraph{The 15-model pool.} All fifteen models are served concurrently as separate vLLM servers with tensor-parallel size 1, each holding a fixed share of one B200 (Table~\ref{tab:scale15}). The pool spans 0.5B to 32B and mixes general, mathematics, code and reasoning specialists, so the executor faces a genuine capability-versus-capacity choice at every step. Serving is \texttt{bfloat16} with an 8192-token context, 2048 output tokens and \texttt{-{}-max-num-seqs 16} on every server, except the two mathematics specialists, which are served at 4096 context and 1024 output tokens. Training uses 300 problems from the MATH training split and evaluation the same 500 held-out test problems as the main results. The replay uses MasRouter's final checkpoint after 5 epochs (learning rate $0.01$, batch size 32, cost-quality coefficient zero) and \sys{}'s final checkpoint after 15 epochs, reused from the preceding pool-scaling run. \sys{} additionally trains the executor across arrival rates $\{10,30,50\}$ and budget tiers $\{10,30,50,100,200,300,600,1000\}$\,s; this is a comparison of the resulting policies, not equal training effort. Its executor entropy coefficient is $0.03$ for the 45-action space, with evaluation temperatures $0.3$ (executor) and $0.7$ (planner). These settings follow the pool-scaling recipe; neither replay checkpoint was selected by a validation sweep. Each request carries a deadline drawn from $\{10, 30, 50, 100, 200, 300\}$\,s by item identifier, so both methods see identical deadlines.

\begin{table}[h]
  \centering
  \small
  \caption{The 15-model pool used for the scalability study, three B200s, tensor-parallel size 1 throughout.}
  \label{tab:scale15}
  \begin{tabular}{cp{0.68\textwidth}c}
    \toprule
    \textbf{GPU} & \textbf{Models served concurrently (memory fraction)} & \textbf{Total} \\
    \midrule
    0 & DeepSeek-R1-Distill-Qwen-7B (0.15), Qwen2.5-Math-7B (0.15), Qwen2.5-Coder-7B (0.15), Llama-3.2-3B (0.08), Qwen2.5-Coder-1.5B (0.05), Qwen2.5-0.5B (0.04) & 0.62 \\
    1 & Llama-3.1-8B (0.16), Qwen2.5-7B (0.15), Phi-3.5-mini (0.09), DeepSeek-R1-Distill-Qwen-1.5B (0.05), Qwen2.5-Math-1.5B (0.05), Llama-3.2-1B (0.04) & 0.54 \\
    2 & Qwen2.5-32B (0.46), Qwen2.5-Coder-14B (0.23), Mistral-Nemo-12B (0.19) & 0.88 \\
    \bottomrule
  \end{tabular}
\end{table}

\paragraph{Arrival-trace replay.} We submit benchmark questions at publicly released Azure LLM inference arrival timings~\citep{patel2024splitwise}, rescaling their mean rate while preserving inter-arrival structure. This evaluates production-derived arrival patterns on our benchmark deployment. The code trace has an inter-arrival coefficient of variation of $13.2$; the conversation trace is near-Poisson, with a coefficient close to $1$. The five-model comparison holds the MATH workload and 60\,s budget fixed while changing arrival timing. The 15-model study uses the per-item budget tiers above.

\section{Paired Executor Ablations}
\label{app:ablation}

\begin{table}[h]
  \centering
  \small
  \caption{Paired executor ablations ($n{=}500$ per arm). MATH uses 50 requests/min and per-item deadlines from $\{10,30,50,100,200,300\}$\,s; MMLU-Pro uses 100 requests/min and a 300\,s deadline. ``Delivered'' is the percentage of all requests both correct and on time. Protocols are identical within each pair.}
  \label{tab:ablation_summary}
  \resizebox{\columnwidth}{!}{%
  \begin{tabular}{llrrr}
    \toprule
    \textbf{Workload} & \textbf{Executor} & \textbf{Acc (\%)} & \textbf{Lat (s)} & \textbf{Delivered (\%)} \\
    \midrule
    MATH & Learned model--strategy selection & 71.4 & 159.7 & 33.8 \\
         & Uniform random model--strategy selection & 65.0 & 102.4 & 30.0 \\
    \midrule
    MMLU-Pro & Full infrastructure inputs & 57.8 & 373 & 28.0 \\
             & Queue, latency and KV inputs zeroed; retrained & 61.0 & 739 & 21.8 \\
    \bottomrule
  \end{tabular}}
\end{table}

The MATH pair retains the same planner and scheduler and changes only executor selection. The MMLU-Pro pair retains semantic and budget inputs and retrains the executor without serving telemetry. Its raw accuracy--latency trade-off shows why delivered accuracy matters: telemetry guides the policy toward computation that finishes within the available time. Deadline attainment is $55.0\%$ for the full executor and $37.4\%$ without telemetry.

\Needspace{16\baselineskip}
\section{Infrastructure Blindness: Full Invariance Measurement}
\label{app:invariance}

\begin{table}[h]
  \centering
  \small
  \caption{Infrastructure blindness as a measurable invariance. Share of all LLM calls MasRouter sends to each model on MATH, as offered load rises $20\times$. A load-aware policy would redistribute; this one does not move.}
  \label{tab:invariance}
  \begin{tabular}{lcccccc c}
    \toprule
    \textbf{Model} & \textbf{10} & \textbf{30} & \textbf{50} & \textbf{100} & \textbf{150} & \textbf{200} & \textbf{max shift} \\
    \midrule
    Qwen2.5-Coder-14B   & 32.9\% & 33.2\% & 33.0\% & 32.2\% & 33.0\% & 33.0\% & 0.92\,pp \\
    DeepSeek-R1-32B     & 18.2\% & 18.2\% & 18.6\% & 18.8\% & 18.6\% & 17.7\% & 1.11\,pp \\
    Mistral-Small-24B   & 25.9\% & 25.9\% & 25.7\% & 25.5\% & 25.5\% & 26.0\% & 0.52\,pp \\
    Llama-3.1-8B        & 16.1\% & 16.1\% & 16.3\% & 17.0\% & 16.5\% & 16.9\% & 0.94\,pp \\
    Llama-3.2-3B        &  6.9\% &  6.6\% &  6.5\% &  6.4\% &  6.5\% &  6.3\% & 0.54\,pp \\
    \bottomrule
  \end{tabular}
\end{table}

\section{Telemetry Staleness}
\label{app:staleness}

The sweep delays each telemetry reading consumed by the executor while leaving the serving layer unchanged. On MMLU-Pro at 100 requests/min, delays of 1--10\,s preserve accuracy and deadline attainment relative to fresh readings (Table~\ref{tab:staleness}). This tests tolerance to delayed observations at the measured load and queue dynamics.

\begin{table}[h]
  \centering
  \small
  \caption{Telemetry staleness sweep, MMLU-Pro at 100 req/min, $n{=}500$ per arm, 300\,s deadline.}
  \label{tab:staleness}
  \begin{tabular}{lcccc}
    \toprule
    \textbf{Delay} & 0\,s & 1\,s & 5\,s & 10\,s \\
    \midrule
    Accuracy (\%)    & 57.8 & 60.4 & 59.2 & 58.8 \\
    Mean latency (s) & 373 & 362 & 351 & 352 \\
    SLO (\%)         & 55.0 & 57.0 & 57.8 & 56.6 \\
    \bottomrule
  \end{tabular}
\end{table}

\section{Statistical Tests}
\label{app:stats}

\paragraph{Paired evidence from the larger pool.} The 15-model study evaluates both methods on the same 500 problems and assigned deadlines. The 95\% Wilson intervals~\citep{wilson1927probable} for accuracy are $[70.4,78.0]\%$ for \sys{} and $[66.0,74.0]\%$ for MasRouter; for deadline attainment they are $[69.8,77.5]\%$ and $[12.0,18.2]\%$; for correct, on-time answers, $[50.4,59.1]\%$ and $[9.1,14.7]\%$. Latency is a continuous measurement and is not assigned a binomial interval.

Continuity-corrected paired McNemar tests~\citep{mcnemar1947note} use the disagreements on each binary outcome. For accuracy, 61 problems are solved only by \sys{} and 40 only by MasRouter ($p{=}0.047$). For deadline attainment the counts are 302 versus 7, and for correct, on-time answers 222 versus 6 (both $p{<}10^{-4}$). These tests quantify variation across paired evaluation items, rather than independent training seeds.

\section{Blackbox and Hybrid Implementation Details}
\label{app:blackboxdetails}

\paragraph{Sliding-window RPM tracking.} We use a 60-second sliding window per model. Each outgoing request is timestamped; the congestion signal is the count of timestamps within the window divided by the model's RPM limit. When this ratio exceeds 1.0, excess requests are held in a local queue and dispatched as capacity frees up, creating backpressure visible to the executor.

\paragraph{Dual Lagrange multipliers.} For hybrid pools, we maintain $\lambda_{\text{time}}$ and $\lambda_{\text{money}}$ with independent dual updates:
\begin{align}
  \lambda_{\text{time}} &\leftarrow \text{clip}(\lambda_{\text{time}} + \eta_\lambda \cdot (\bar{C}_{\text{time}} - 1),\; 0,\; \lambda_{\max}) \\
  \lambda_{\text{money}} &\leftarrow \text{clip}(\lambda_{\text{money}} + \eta_\lambda \cdot (\bar{C}_{\text{money}} - 1),\; 0,\; \lambda_{\max})
\end{align}
where $\bar{C}_{\text{time}} = \sum \ell_k / \beta_{\text{time}}$ and $\bar{C}_{\text{money}} = \sum c_k / \beta_{\text{money}}$. With $\beta_{\text{money}} = \infty$ (no monetary budget) and no blackbox models, $\lambda_{\text{money}}$ stays at zero and the formulation reduces exactly to the whitebox-only system in the main text.

\end{document}